\documentclass[sigconf]{acmart}
\usepackage{blindtext}
\usepackage{graphicx}
\usepackage{subcaption}
\usepackage[linesnumbered,ruled]{algorithm2e}
\usepackage{color}
\usepackage{xcolor}
\usepackage{booktabs}
\usepackage{multirow}

\usepackage{fancyhdr}
\AtBeginDocument{%
  \providecommand\BibTeX{{%
    \normalfont B\kern-0.5em{\scshape i\kern-0.25em b}\kern-0.8em\TeX}}}

\copyrightyear{2021}
\acmYear{2021}
\setcopyright{acmcopyright}\acmConference[SEC '21]{The Sixth ACM/IEEE Symposium on Edge Computing}{December 14--17, 2021}{San Jose, CA, USA}
\acmBooktitle{The Sixth ACM/IEEE Symposium on Edge Computing (SEC '21), December 14--17, 2021, San Jose, CA, USA}
\acmPrice{15.00}
\acmDOI{10.1145/3453142.3491290}
\acmISBN{978-1-4503-8390-5/21/12}

\renewcommand\footnotetextcopyrightpermission[1]{} % removes footnote with conference information in first column
\begin{document}
% The below commands remove default conference info from the draft manuscript

\title{BitTrain: Sparse Bitmap Compression for Memory-Efficient Training on the Edge}


\author{Abdelrahman Hosny} 
%\stepcounter{footnote}
\authornote{Both authors contributed equally to this research.}

\affiliation{%
  \institution{Department of Computer Science \\ Brown University}
  \city{Providence}
  \state{RI}
  \country{USA}
  \postcode{02912}
}
\email{abdelrahman_hosny@brown.edu}

\author{Marina Neseem}
\authornotemark[1]

\affiliation{%
  \institution{School of Engineering \\ Brown University}
  \city{Providence}
  \state{RI}
  \postcode{02912}
  \country{USA}
}
\email{marina_neseem@brown.edu}

\author{Sherief Reda}
\affiliation{%
  \institution{School of Engineering \\ Brown University}
  \city{Providence}
  \state{RI}
  \postcode{02912}
  \country{USA}
}
\email{sherief_reda@brown.edu}

\renewcommand{\shortauthors}{Hosny and Neseem, et al.}

\begin{abstract}
    Training on the Edge enables neural networks to learn continuously from new data after deployment on memory-constrained edge devices. Previous work is mostly concerned with reducing the number of model parameters which is only beneficial for inference. However, memory footprint from activations is the main bottleneck for training on the edge. Existing incremental training methods fine-tune the last few layers sacrificing accuracy gains from re-training the whole model. In this work, we investigate the memory footprint of training deep learning models, and use our observations to propose BitTrain. In BitTrain, we exploit activation sparsity and propose a novel bitmap compression technique that reduces the memory footprint during training. We save the activations in our proposed bitmap compression format during the forward pass of the training, and restore them during the backward pass for the optimizer computations. The proposed method can be integrated seamlessly in the computation graph of modern deep learning frameworks. Our implementation is safe by construction, and has no negative impact on the accuracy of model training. Experimental results show up to 34\% reduction in the memory footprint at a sparsity level of 50\%. Further pruning during training results in more than 70\% sparsity, which can lead to up to 56\% reduction in memory footprint. BitTrain advances the efforts towards bringing more machine learning capabilities to edge devices. Our source code is available at \url{https://github.com/scale-lab/BitTrain}.
\end{abstract}

%   <concept>
%       <concept_id>10010147.10010257.10010293.10010294</concept_id>
%       <concept_desc>Computing methodologies~Neural networks</concept_desc>
%       <concept_significance>500</concept_significance>
%       </concept>
%   <concept>
%       <concept_id>10003120.10003138.10011767</concept_id>
%       <concept_desc>Human-centered computing~Empirical studies in ubiquitous and mobile computing</concept_desc>
%       <concept_significance>500</concept_significance>
%       </concept>
%  </ccs2012>

% \ccsdesc[500]{Computing methodologies~Neural networks}
% \ccsdesc[500]{Human-centered computing~Empirical studies in ubiquitous and mobile computing}

\keywords{Training on the Edge, Memory Footprint, Edge Intelligence, Resource Constrained Devices}

\pagestyle{fancy}
\fancyhf{}
\lhead{To appear in the proceedings of The Sixth ACM/IEEE Symposium on Edge Computing (SEC 2021)} % clear all header fields

\fancyfoot{} % clear all footer fields

\maketitle

\thispagestyle{fancy}
\cfoot{\thepage}
\section{Introduction}
\label{sec_into}
Over the past decade, deep learning has achieved unprecedented successes in various domains.
Researchers have realized the benefits of deploying deep learning models on edge devices; therefore, they started to develop techniques to make them more resource efficient \citep{han2016deep}. 
However, only deploying the pre-trained models on edge devices is not sufficient. 
Edge devices are continuously collecting rich and sensitive data.
This new data can be used to fine-tune those models which would significantly improve their performance and their adaptive capability to new environments.

A prime example for on-device learning is model personalization.
With advances in digital services, large tech companies strive to make their services as unique to each of their users as possible.
For example, personal assistants such as Siri (Apple), Alexa (Amazon), Cortana (Microsoft), and Google Assistant recognize the voice and the accent of their owner, and learn to not recognize other voices after their initial setup.
Human activity recognition \cite{lin2020model}, health applications \cite{suo2018deep} and smart home appliances \cite{jin2019personalized} also demand model personalization to improve user satisfaction. 
Model personalization defeats the purposes of \textit{generalizability}, which is the primary metric for the performance of deep learning models.
Training on the edge not only makes model personalization more feasible, but also keeps data private and safe.

\begin{figure*}[t]
\centering

\begin{tabular}{c}

\subfloat[\label{memory_w_batch_size}]{\includegraphics[width=0.33\textwidth]{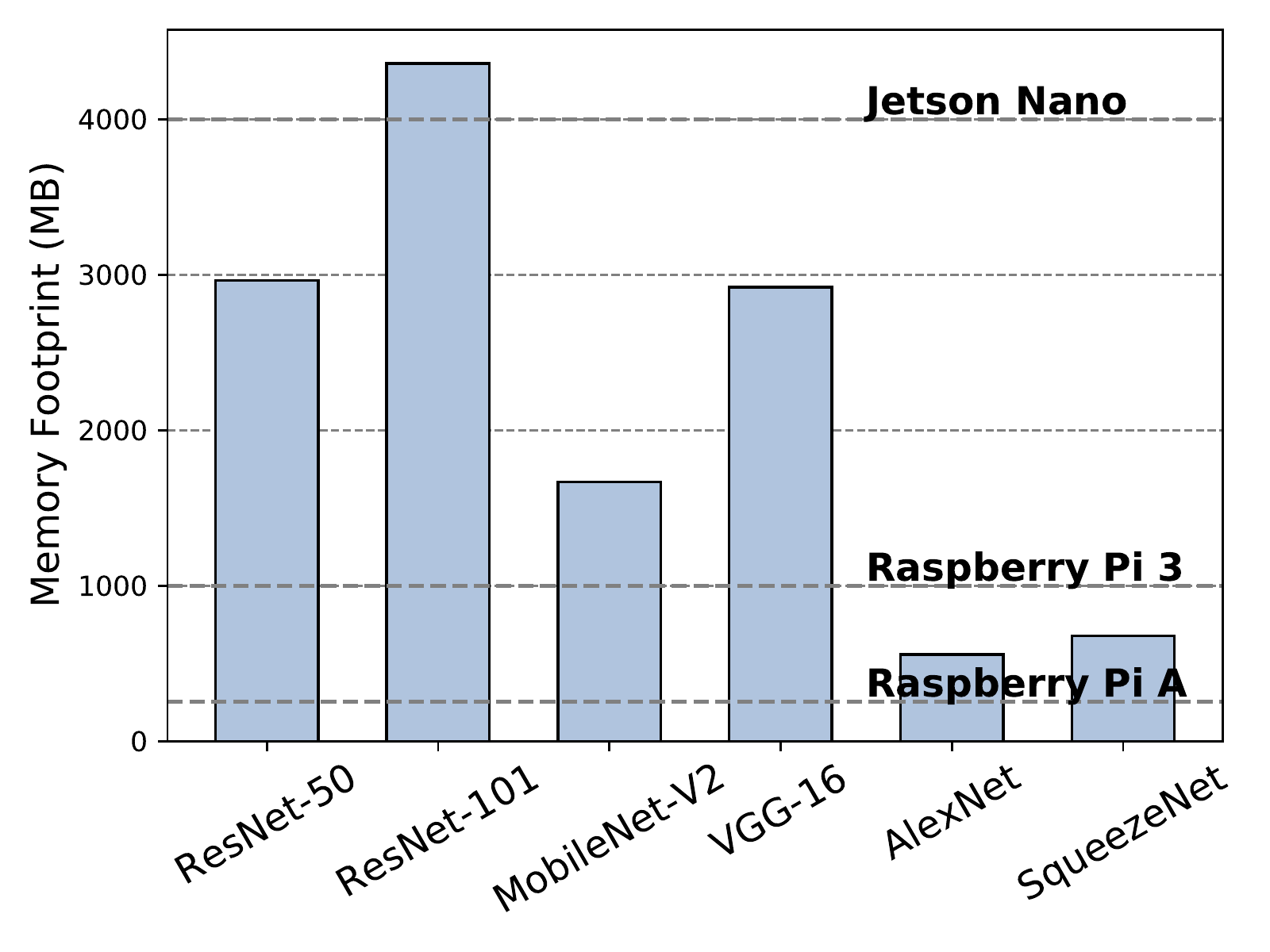}}

\subfloat[\label{memory_components_different_models}]{\includegraphics[width=0.33\textwidth]{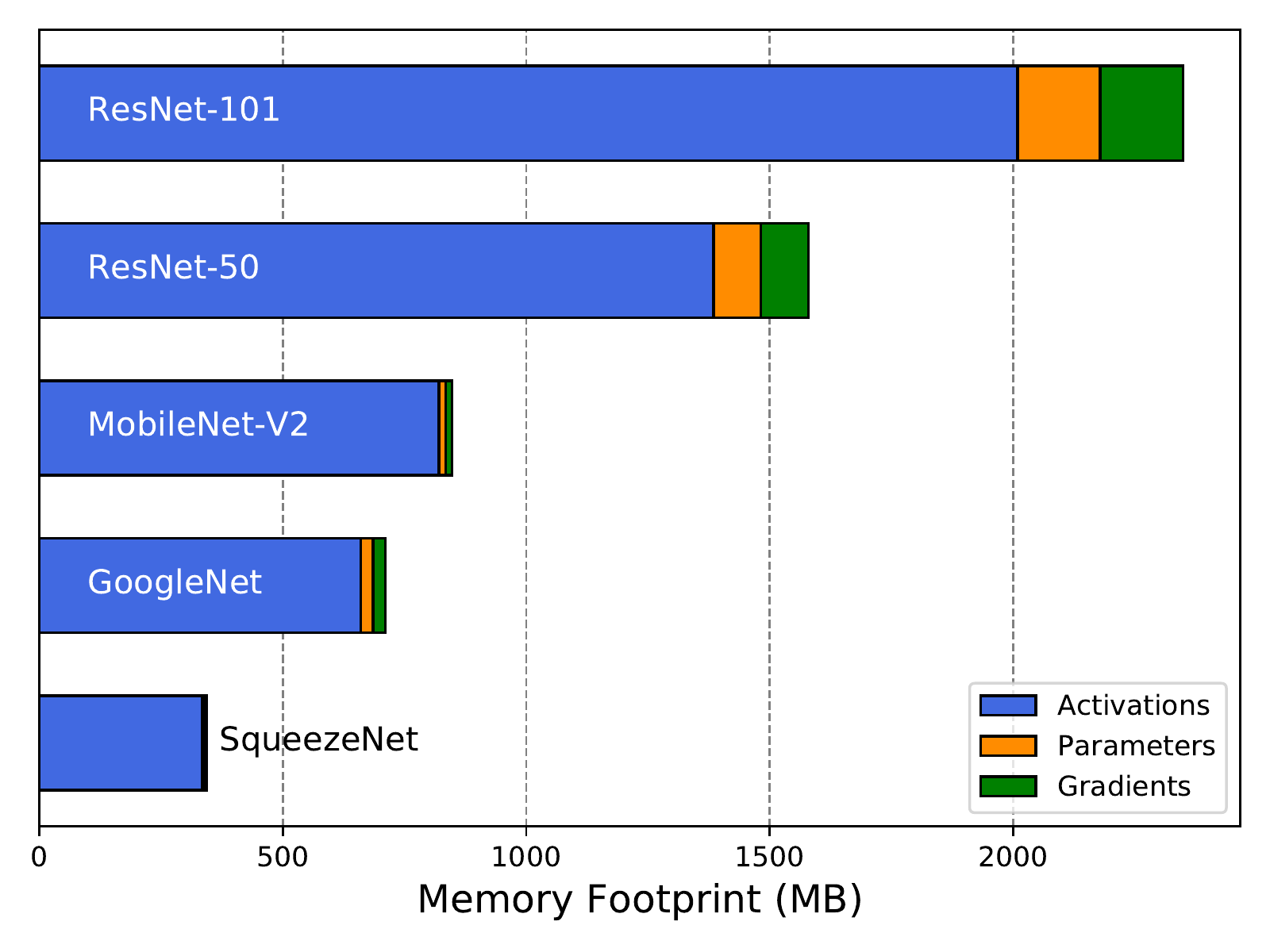}}

\subfloat[\label{memory_w_batch_size}]{\includegraphics[width=0.33\textwidth]{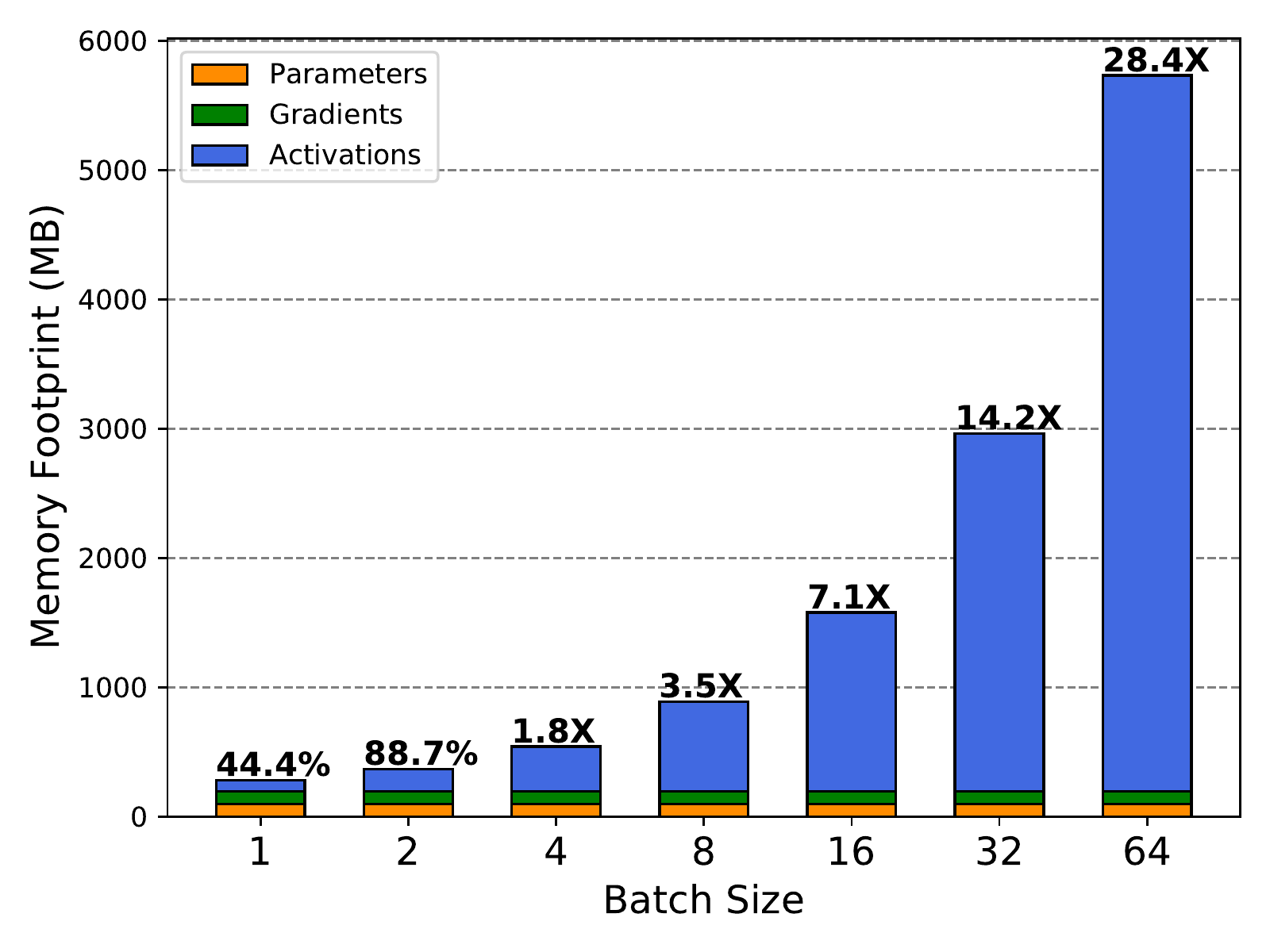}}

% \subfloat[Activations values histogram for ResNet\label{activations_histogram}]{\includegraphics[width=0.45\textwidth]{figs/resnet101_activations_hist.pdf}}

\end{tabular}
\belowcaptionskip -10 pt
\caption{Memory Footprint. (a) Training Memory Footprint at Batch Size 32. (b) Memory Footprint Components for MobileNet-v2, ResNet-50, and Resnet-101 with Batch Size 16. (c) Memory footprint using different batch sizes for ResNet-50.}
\label{fig:theoretical_analysis_for_intro}
\end{figure*}

Current approaches send the data to cloud servers in order to execute training epochs to fine-tune the models.
After that, updated versions of the models are deployed on the edge devices.
However, this approach risks the privacy of the data which could be sensitive, such as medical data that are protected by HIPAA regulations \citep{vepakomma2018split}.
Moreover, continuously syncing data requires a huge network bandwidth.
For example, traffic surveillance cameras deployed at every street intersection in a city would require sending gigabytes of data to the cloud everyday, which is extremely expensive.
Furthermore, it can be even unfeasible due to weak or limited internet connection as it is the case in remote agricultural lands \cite{leaning-iot}, or even in space exploration missions (e.g. Mars Rovers) \citep{squyres2003athena, volpe1996rocky}.
That is the reason why on-device learning is essential to push the limits of edge capabilities.

On-device learning is significantly challenging due to the energy, compute, and memory constraints on the edge devices.
Some work has started exploring training on the edge \citep{han2016deep, dynamic}.
Memory footprint is one of the main challenges for training on the edge.
Figure \ref{fig:theoretical_analysis_for_intro}-a shows the memory required for training some of the modern computer vision models.
We observe that even the memory of a Jetson Nano board is insufficient for training an average state-of-the-art model \cite{suzen2020benchmark}.
During training, memory has three main components: (i) the model parameters, (ii) the activations for each layer computed during the forward pass, and (iii) the gradients computed during the backward pass. 
In Figure \ref{fig:theoretical_analysis_for_intro}-(b), we see that the memory for activations is the dominant factor. 
In addition, the memory footprint increases significantly as the batch size increases as shown in Figure \ref{fig:theoretical_analysis_for_intro}-(c).
However, little work has been devoted in optimizing the activations memory with respect to the amount of work invested in optimizing the parameters memory.
The main reason is that model optimization has been the main use case for deploying models for inference on the edge devices.
To solve this problem, \citet{inthewild} proposed retraining the fully connected layers only, while Cai \textit{et. al.} \cite{tinytl} suggest training the biases and the fully connected layers only, while freezing all the weights. 
In other words, the idea is to not save the activations because they are only needed to compute the weights gradients, which is partially discarded in their proposals.
However, this limits the network's capacity to learn from the new data, and reduces the accuracy gains from retraining. 
Having the flexibility to tune all the weights of a model maximizes the benefits of on-device learning.

Other researchers have explored general techniques to reduce the memory footprint of training regardless of the used hardware. 
Model parameters can be sparsified throughout training, which reduces the number of model parameters and gradients, leaving the activations memory unaffected \cite{sparsetraining}.
Using half precision \cite{mixedprecision} and reducing the batch size \cite{microbatching} have a direct impact on reducing the memory footprint of training, and improve parallelization.
However, these techniques introduce a toll on the accuracy of the pre-trained models.
Furthermore, checkpointing reduces the memory footprint during training by only storing the activations of a subset of layers, and recomputing the needed layers again during backpropagation \cite{checkpointing}.
This provides a trade-off between the memory footprint and the number of floating-point operations (FLOPs).
Nonetheless, this method is targetted towards training deeper models on server-scale GPUs.

In this work, our goal is to reduce the memory footprint of model training on the edge without affecting the accuracy.
The rationale is that we opt for training on the edge to improve the accuracy of the already trained models, while complying with constraints (e.g. data privacy, cloud connectivity, bandwidth).
That raises a fundamental question ``How much memory is needed for training?'', and more specifically ``How much memory is needed to store the activations?''.
By analyzing the activations, we found that activations by nature are sparse.
More than 70\% of the stored activations are zeros due to ReLU non-linearity which is used in most neural network models.
We leverage this observation to make on-device learning more feasible.

In BitTrain, we propose to detach the activation storage from their involvement in computations. This allows us to compress the activations for later use during the backward pass.
We also introduce activations pruning which can further increase the memory savings while producing a memory accuracy trade-off. 
Our contributions can be summarized as follows:
\begin{itemize}
    \item In modern deep learning frameworks, we detach the activations storage from the \textit{Tensor} representation in computation graphs\footnote{Computation graphs are how modern deep learning frameworks represent neural networks for both training and inference.}, allowing us to address the memory footprint issues of neural network activations.
    \item We present BitTrain, a novel Bitmap Sparse Compression method to efficiently store the activations with negligible computational overhead, and with no change to the underlying computation graph. By construction, our compression is safe and has no negative impact on the model accuracy.
    \item We analyze the theoretical and empirical memory reduction by using our method. Experimental results show that we can achieve up to 34\% memory saving at a sparsity level of 50\% per convolution activation. Combining our method with existing work that increase sparsity, we can achieve up to 56\% memory saving at a sparsity level of 75\%.
    \item Since BitTrain is orthogonal to existing methods, we study the effect of combining our method with existing techniques, namely: low-precision training, activation pruning, and checkpointing. Then, we discuss how each combination affects the amount of memory required for training.
    
\end{itemize}

The rest of the paper is organized as follows. 
In Section \ref{sec_relatedwork}, we give a brief background on the most relevant use cases for on-device learning, and establish a foundational framework for advancing the state-of-the-art in on-device training.
Using our framework, we analyze existing methodologies and discuss how they relate to the broader goals of training on the edge.
Then, we present our methodology for reducing the memory footprint in Section \ref{sec_methodology}. 
In Section \ref{sec_experiments}, we present a detailed theoretical and empirical analysis of our method.
Moreover, we investigate the gains from combining our method with existing techniques.
Finally, we conclude our study in Section \ref{sec_concolusion}.

\section{Background and Related Work}
\label{sec_relatedwork}
\begin{figure}[t!]
    \centering
    \includegraphics[scale=0.5]{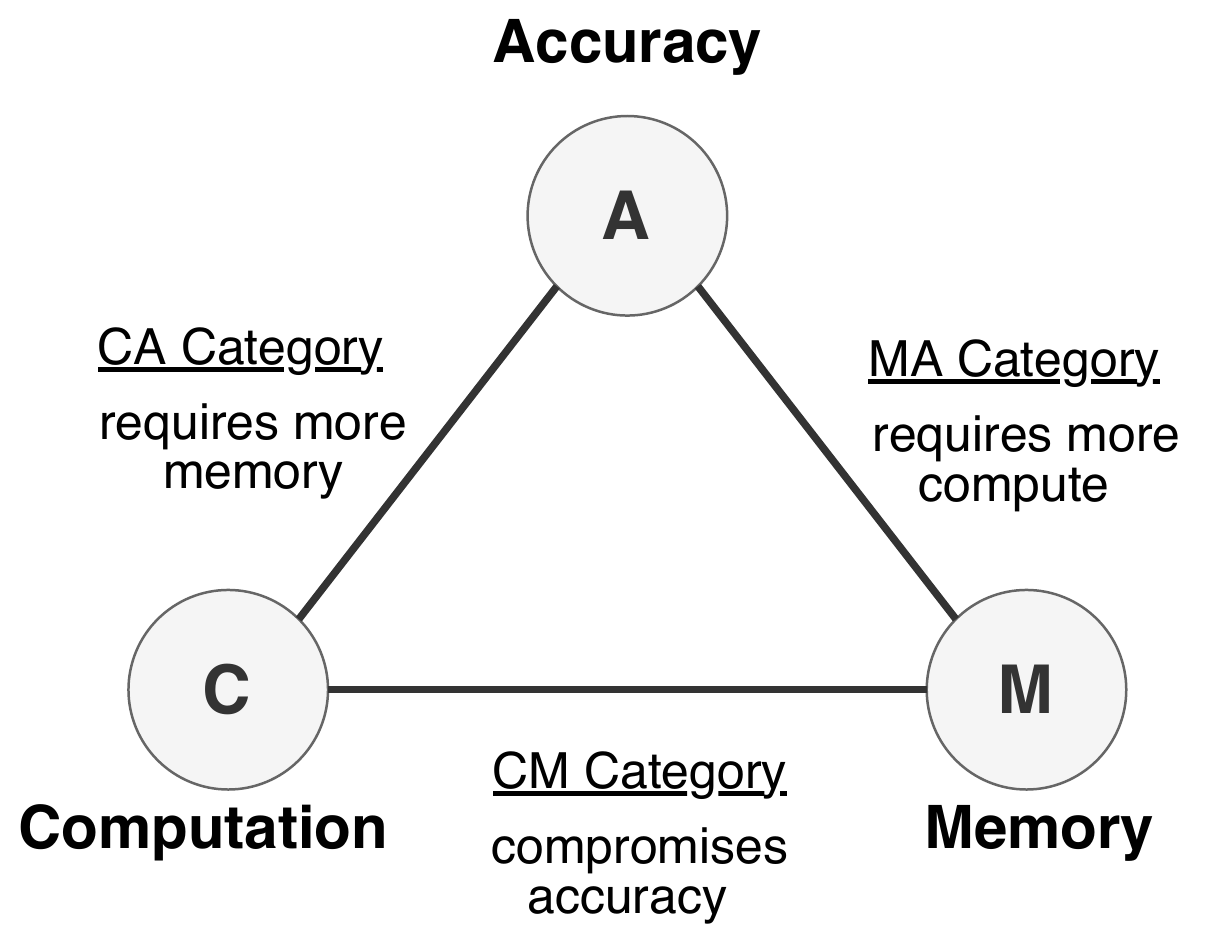}
    \belowcaptionskip -20 pt
    \caption{Compute, memory and accuracy trade-off in training on the edge.}
    \label{fig:related-work-framework}
\end{figure}

\textbf{Transfer Learning.}
Deep learning models trained on large datasets (e.g. ImageNet \cite{deng2009imagenet}) can be widely used to retrain neural networks on the edge with local data.
The idea is to keep the parameters of the feature extraction layers unchanged, and only train the last layers \cite{pan2009survey}.
Transfer learning on the edge can be used for customization of mobile services as well as for offline retraining.
It also serves federated learning, where a number of edge devices collaborate to learn a shared model \cite{mcmahan2017communication}.
This approach saves training memory because the intermediate activations for the feature extractor do not need to be stored.
However, the accuracy can significantly drop, especially when the new data is coming from a distribution that is very far from the distribution of the data used during the initial training.
To solve this issue, Cai \textit{et al.} \cite{tinytl} proposed fine-tuning the both the final layers as well as the biases of the feature extractor (i.e intermediate activations are not needed to compute the gradients for the biases). This approach saves the memory footprint; however, fine-tuning all the layers significantly increase the ability of the model to adapt on the new data.

\textbf{Compressing Models for Inference. }
Over the past few years, researchers have been continuously advancing the state of the art in neural networks with deeper models that result in higher accuracy when deployed.
This made the training and the deployment of these models both compute- and memory-intensive.
Optimizing for the compute and memory resources has been a concern for model inference on the edge \cite{chen2019deep}.
Pruning \citep{blalock2020state, janowsky1989pruning, yu2018nisp, suzuki2018spectral} and quantization \citep{hashemi2017understanding, guo2018survey, lin2016fixed} are on the top of the most widely adopted techniques for model compression.
The idea is to reduce the number of parameters in the model in order to fit in less memory and use less compute.
However, these methods are designed mainly for inference and do not consider the memory footprint of model training.
Another approach is to design handcrafted models that are small enough to fit on edge devices \citep{howard2017mobilenets, tan2020efficientdet}. 
Although these models are compressed into less parameters, the memory footprint needed during training for storing the activations is still high.
As their current design stand, it is infeasible to adopt these techniques alone to reduce the memory footprint of model training on the edge as they require a large pre-trained dense model.

\textbf{Training Memory.}
Training on the edge is challenging, and in order to achieve real-world adoption, a number of factors need to be considered.
In Figure \ref{fig:related-work-framework}, we look at the effort done in this direction through a hypothesized three-axes framework; namely: Computation, Accuracy and Memory.
We observe that the state-of-the-art work optimizing any two of these desired characteristics inversely affects the third one.
As we will discuss next, techniques that minimize both computation and memory incurs accuracy loss.
These techniques would be suitable for some use-cases where network-connectivity is unreliable, and training is done from scratch.
Techniques that maintain high accuracy of server-grade training are desirable for other uses-cases that such as transfer learning and model personalization.
However, they either compromise on the memory or the amount of computation needed.
In the following we look at different techniques in the literature through the lens of this framework, and then describe how our work capitalizes on all of these efforts.

\textbf{Low Precision (CM Category).} 
Micikevicius \textit{et al.} \citep{mixedprecision} use half precision (16 bits) for weights, gradients, and activations.
This reduces the memory footprint by a factor of $2 \times$, and it can be complementary to any other low-memory training technique to maximize the savings.
Courbariaux \textit{et al.} \citep{courbariaux2014training} show that they can train models using 10-bits multiplications without severely affecting the accuracy.
Jia \textit{et al.} \citep{jia2018highly} increase the training throughput of a single GPU using a mixed-precision training method.
Dipankar \textit{et al.} \citep{das2018mixed} use fixed-point integer operations to train the models.
They show that this can achieve competitive results to training with floating-point operations.
All of these techniques use lower precision to reduce the memory footprint of training and possibly the number of operations needed, which compromises on the accuracy of the model.

\textbf{Microbatching (CM Category).}
Huang \textit{et al.} \cite{microbatching} use microbatch-based training where they can sequentially send smaller subsets of the batch through the network, and accumulate the gradients until the whole batch is processed.
Then, gradient update is executed once.
This approach reduces the memory footprint without affecting the total number of operations performed.
It is important to note that microbatching has a direct impact on the statistical characteristics of batch normalization layers.
That is why it needs to be exercised carefully in order to avoid losing accuracy.

\textbf{Sparsity (CA Category).} 
A sparse matrix is one in which most of the values are zero.
In general, the idea of exploiting sparsity is that multiplications engaging zero elements do not need to be executed, which can be used to reduce the amount of compute.
Pruning is a prime example of exploiting this characteristics, though used for inference model compression.
In model training, techniques that maintain sparsity throughout the entire training process have recently emerged.
For example, Mocanu \textit{et al.} \cite{mocanu2018scalable} introduced a sparse evolutionary training technique reducing quadratically the number of parameters with no effect on accuracy.
Mostafa \textit{et al.} \citep{sparsetraining} proposed a modified version of the algorithm that initializes the network with a fixed sparsity pattern and prune the parameters consequently.
This technique reduces the memory used by the parameters and the gradients only. However, as shown in Figure \ref{fig:theoretical_analysis_for_intro} the parameters and the gradients account for the smallest portion of the memory footprint.
Sparsity has also been exploited to accelerate the training both at the software \cite{spmm_tiling} and the hardware level \cite{tensordash}.
However, these techniques do not reduce the memory footprint.

For highly sparse matrices, storing the non-zero elements and their indices is more efficient than storing all the values in a dense format. 
The most popular format is the Coordinate list format (COO).
It stores each non-zero value (floating-point) along with its n-dimensional indices (fixed-point).
Modern deep learning frameworks offer efficient implementations for the COO format.
However, this format is optimized for reducing the number of matrix operations, and is only memory-efficient if the matrix has a sparsity ratio higher than 80\% (discussed in details in the next section); otherwise, the dense format would consume less memory.

\textbf{Checkpointing (MA Category).}
Chen \textit{et al.} \cite{checkpointing} first proposed the idea of trading computation for memory.
The idea is to discard saving the activations and recalculate them, layer-by-layer, upon backpropagation.
Gruslys \textit{et al.} \cite{checkpointing2} proposed a dynamic programming approach that balances between caching of intermediate results and re-computation.
The interested reader is referred to \cite{lowmemory} for a detailed technical report on combining some of the above techniques for training. 

Our proposed methodology focuses mainly on addressing the memory footprint issue without affecting the accuracy. 
In Section \ref{sec_experiments}, we show that our work is orthogonal to these techniques, and can be combined to further minimize the memory footprint.

\section{Methodology}
\label{sec_methodology}
Deep Learning frameworks represent activations for Convolutional Neural Networks as four-dimensional matrices. 
In modern deep learning frameworks, they are represented in a dense matrix format of dimensions ($batch\_size$, $num\_channels$, $height$, $width$) where $batch\_size$ is the batch size used during either training or testing, $num\_channels$ is the number of channels at any given layer, $height$ and $width$ are the height and the width of the activation maps respectively.
The goal of BitTrain is to reduce the total memory footprint required for training a model.
In the following, we derive the foundations of our method and describe the implementation details.

\begin{figure}[t]
\centering

\begin{tabular}{c}
\subfloat[ResNet-50 \label{resnet_memory_components}]{\includegraphics[width=0.4\textwidth, keepaspectratio]{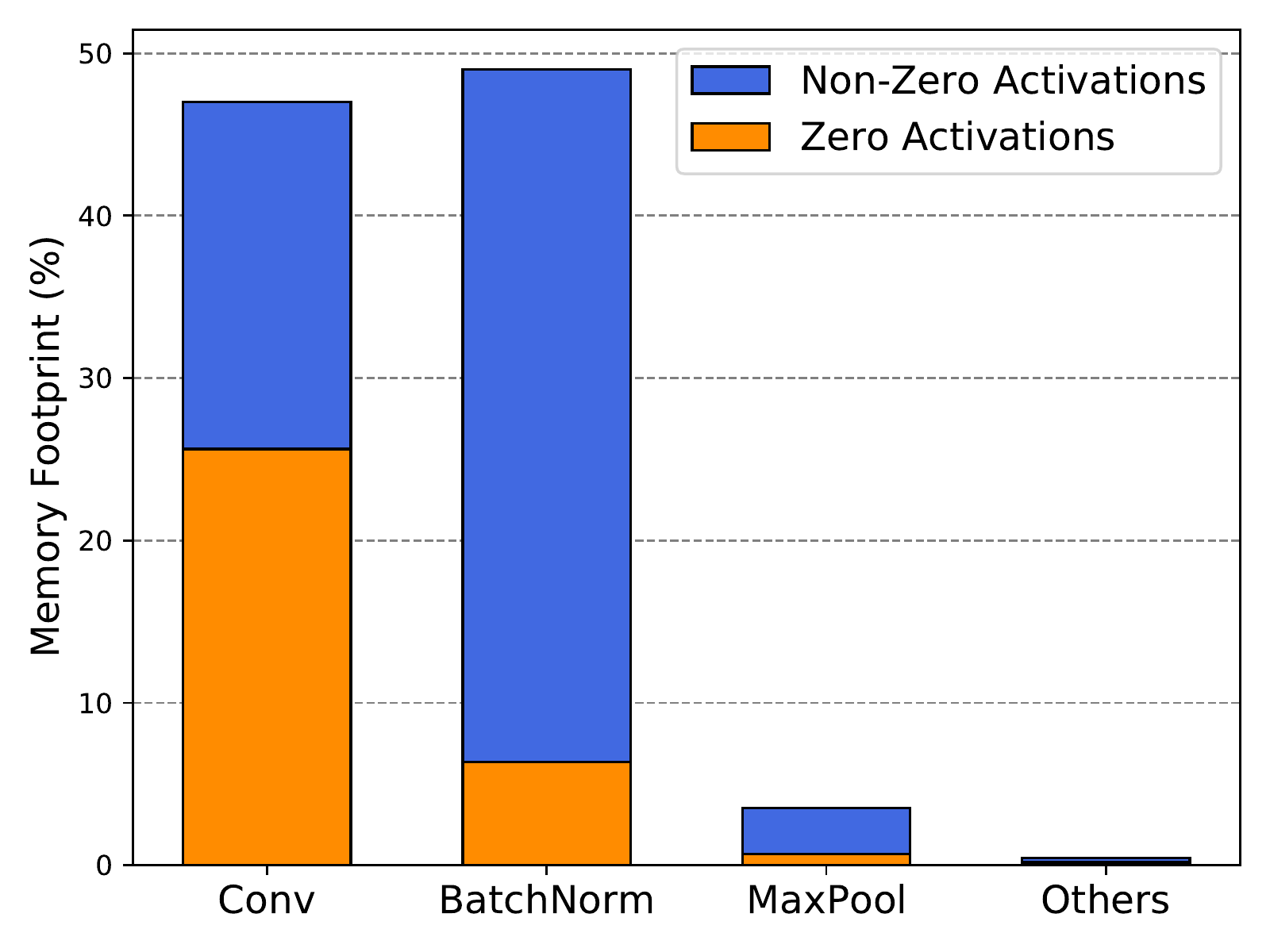}} \\

\subfloat[AlexNet \label{alexnet_memory_components}]{\includegraphics[width=0.4\textwidth, keepaspectratio]{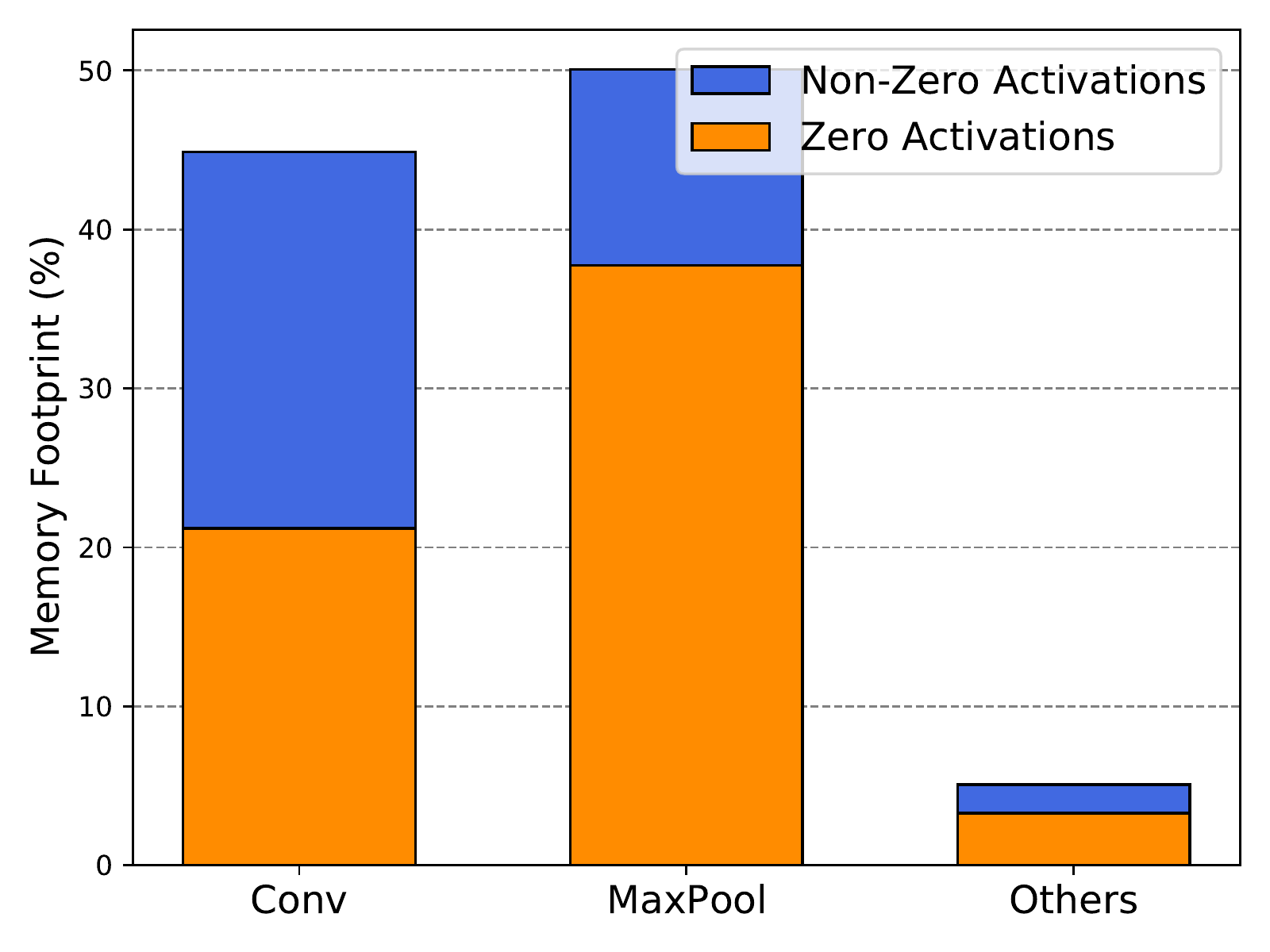}}

\end{tabular}
\caption{Activations Memory footprint for different layers in ResNet-50 and AlexNet.}
\label{activations_per_layer_memory_analysis}
\end{figure}

\subsection{Activation Sparsity in Neural Networks}
\textbf{Exploiting Sparsity.} In the modern deep learning frameworks (i.e. PyTorch and TensorFlow), the input activation for each layer is stored during the forward pass, then those activations are used for calculating the gradients during the backward pass.
Storing the activations for all the layers creates a huge memory footprint during the training process as illustrated earlier in Figure \ref{memory_components_different_models}.
However, most of modern deep learning models use the rectified linear activation function (ReLU) \cite{relu}. 
Using ReLU activation functions results in sparse activations in the successive layers.
We can leverage this sparsity to compress the activations, and hence reduce the memory footprint for training.

In Figure \ref{activations_per_layer_memory_analysis}, we analyze the contribution of the activations of various layers types (e.g. Convolution, Batch Normalization, Max Pooling) to the memory footprint, as well as the average activation sparsity for those layer types throughout the whole models. 
Figures \ref{resnet_memory_components} and \ref{alexnet_memory_components} shows that the memory used to store the activations of the different layers types in ResNet-50 and AlexNet respectively.
For ResNet-50, we notice that the memory used to store the convolution and the batch normalization activations dominates the memory used by the other layers. 
While for AlexNet, the convolution and the max pooling layers activations dominate the memory. 
That means that we should direct our efforts towards reducing the memory used to store the activations of those three layers (convolution, batch normalization, max pooling).

\begin{figure}[t!]
    \centering
    \includegraphics[width=0.45\textwidth]{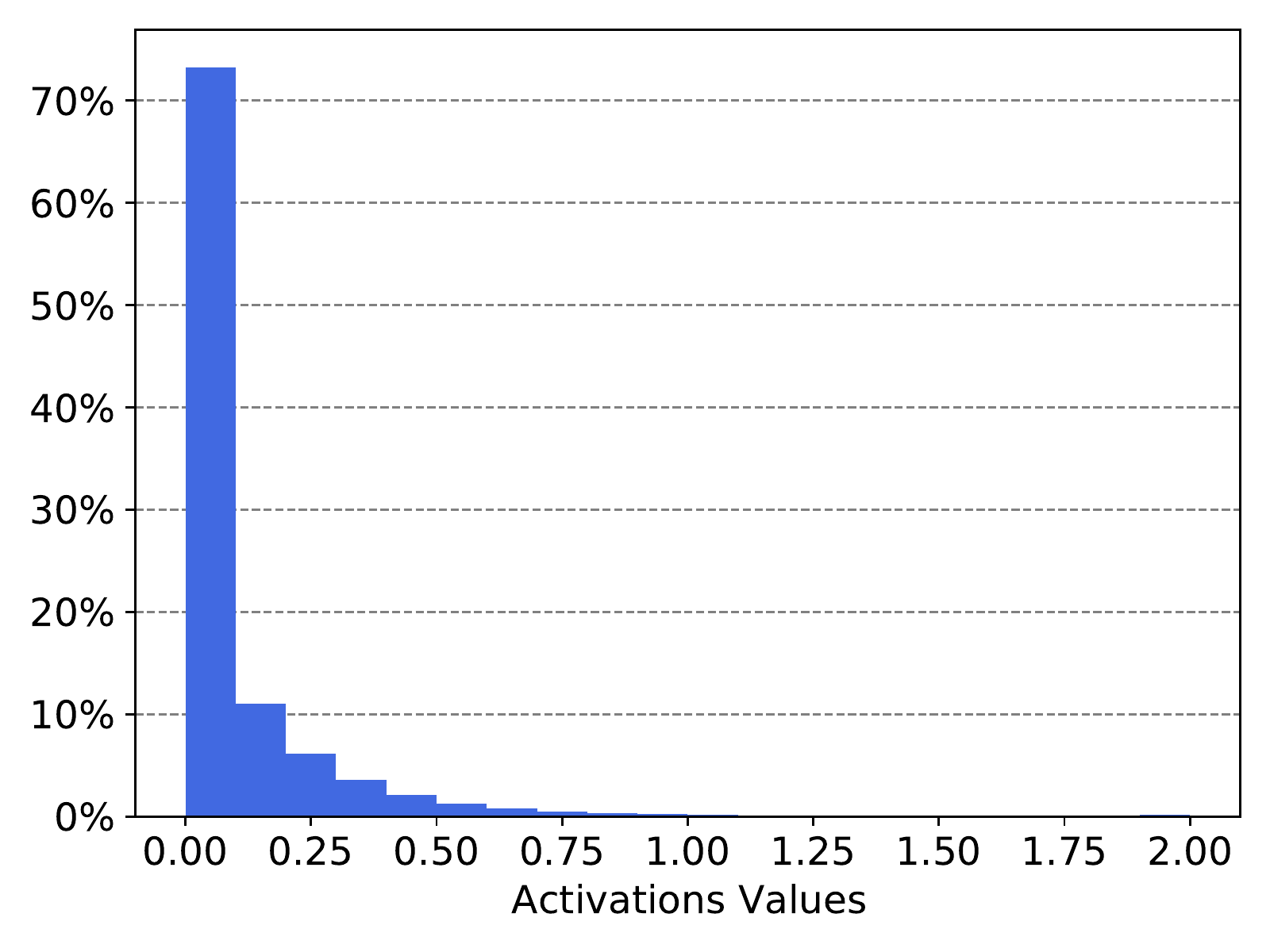}
    \belowcaptionskip -10 pt
    \caption{Activations values histogram for ResNet.}
    \label{fig:activations-histogram}
\end{figure}

From Figure \ref{activations_per_layer_memory_analysis}, we can also notice that the activations sparsity is relatively high for the convolution and the max pooling layers. 
However, it is not a surprise that the activations sparsity for the batch normalization layer is very low. 
That means that using a sparse representation for the batch normalization activations will not offer any reduction in the memory footprint. 
To overcome this problem, we apply the double mask batch normalization introduced by Lieu \textit{et al.} \cite{dynamic}.
The idea is to use the sparse pattern of the the inputs to the normalization layer as a mask to apply for its outputs.
In other words, it propagates sparsity through the batch normalization layer and helps our method reduce more memory.

Moreover, we analyze the activation values throughout the whole network. 
Figure \ref{fig:activations-histogram} shows a histogram of the activation values for all the layers of a pre-trained ResNet model. 
We notice that more that 70\% of the activations are close to zero. 
This implies that we can neglect storing those activations (i.e. we can assume that they are zeros), and hence increase the activations sparsity which would further increase the gain from our compression methodology.

\textbf{Memory vs. Matrix Operations.}
In server-grade model training, a large GPU memory can accommodate model parameters as well as activations calculated during the forward pass.
On edge devices, memory is not only a scarce resource, but may also be shared between the main CPU and the GPU (if exists).
For example, the Nvidia Jetson Nano board houses an ARM Cortex-A57 MPCore processor that shares a 4GB memory with a Maxwell-based GPU that performs up to 4 floating-point operations per clock cycle \cite{arm_isa}.
External memory access has an interrupt latency of at least 200 clock cycles assuming zero wait state \cite{arm_memory_latency} \--- a figure that is empirically higher depending on the system load and the cache status.
Due to the limited memory available, convolution activations from earlier layers will be offloaded to disk (using virtual memory pages), since they are the least recently used.
If no swap memory is available, the training process will be killed by the operating system.

Building on the research done on checkpointing (where activations are not saved at all; instead re-calculated), we propose to trade expensive matrix multiplications for cheaper memory operations.
This trade-off is analogous to \textit{Memoization} in algorithmic contexts \cite{bellman1957dynamic}.
In essence, we take advantage of sparsity and save input activations in a compressed format that leaves more memory for the following dense operations to be performed.
Although the compression and decompression processes add operations to the training loop, they save the disk access time resulting from memory swapping.

\begin{figure}[t!]
    \centering
    \includegraphics[scale=0.55]{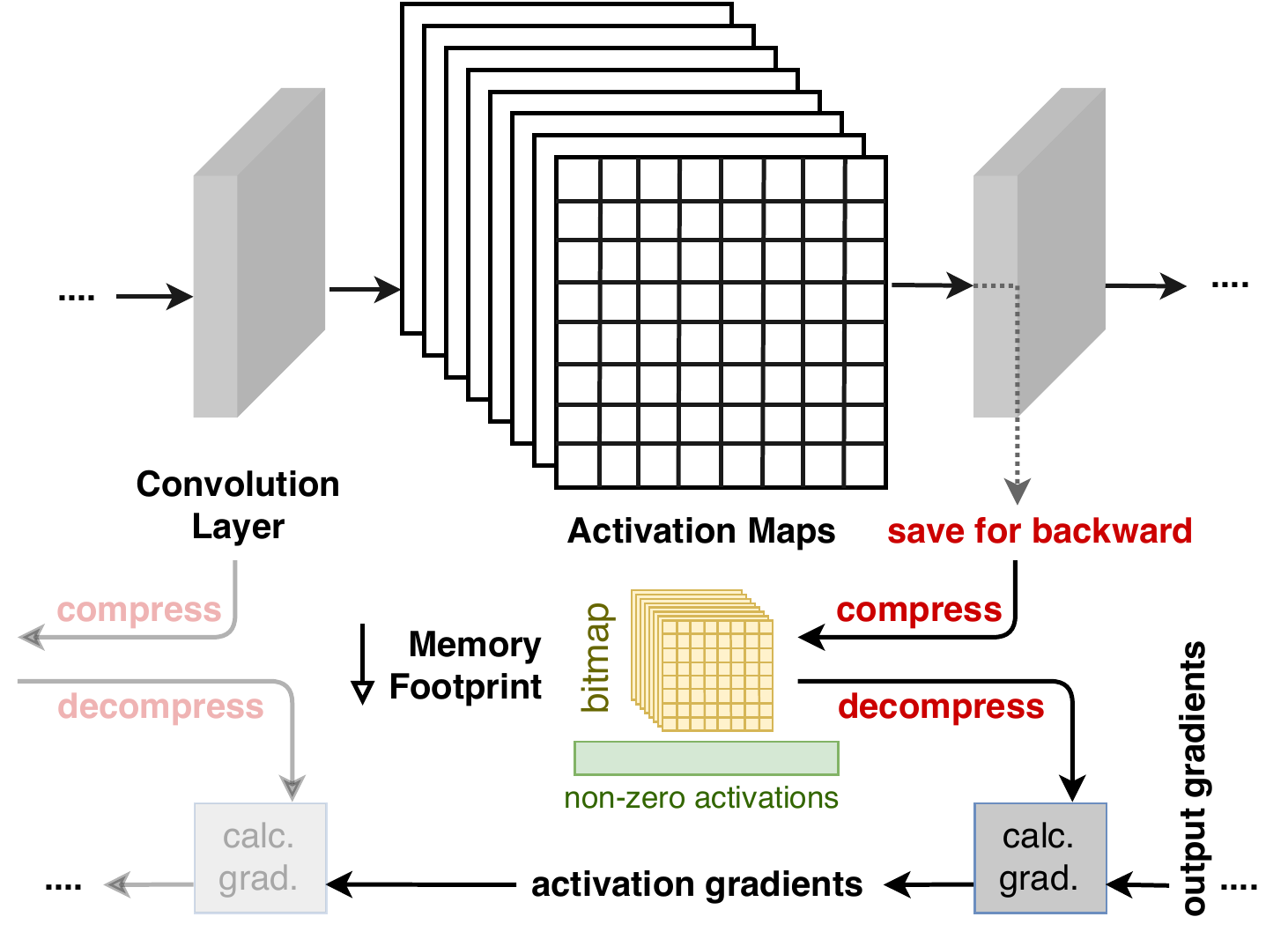}
    \belowcaptionskip -20 pt
    \caption{Compressing dense activations in a bitmap.}
    \label{fig:bitmap-format}
\end{figure}

\subsection{Sparse Bitmap Format}
\textbf{Computation Graphs.}
Modern deep learning frameworks (e.g. Tensorflow \cite{abadi2016tensorflow} and PyTorch \cite{paszke2019pytorch}) have offered an adequate level of abstraction for training deep learning models.
A developer can now imperatively describe the architecture of their neural network, and the framework takes care of the compiling code to lower-level constructs that work efficiently with different hardware interfaces (especially GPUs).
To make this happen seamlessly, these framework construct \textit{a computation graph} that can be used to track and execute the necessary elements of the backpropagation algorithm (in a process called auto-differentiation \cite{paszke2017automatic}).
The computation graph can be either static or dynamic.
In a static computation graph, the neural network is constructed once in the beginning, and then gets attached to a training session.
In this case, memory occupied by the sizes of its tensors (i.e. matrices) is reserved in the beginning. 
On the other hand, dynamic computation graphs get built dynamically, reserving memory for tensors immediately after declaring them, and releasing them when they go out of scope.
This distinction is important in our work, since there is no memory management APIs offered by these frameworks in their Python interfaces.
Optimizing memory has to be implemented at the lower level (using C++), which we describe later in Section \ref{sec_experiments}.

\textbf{Bitmap Format. } 
As discussed in Section \ref{sec_relatedwork}, there is a plethora of sparse matrix representation formats. 
These formats are mainly designed for both storage and operations.
In other words, mathematical operations such as multiplication, division, and inverse are defined and computationally efficient.
We adopt a bitmap format that is optimal for storage as shown in Figure \ref{fig:bitmap-format}. 
During the forward pass, input activations for a given layer is compressed into: (i) a vector containing the non-zero elements, and (ii) a bitmap that sets a bit to 1 at the indices of the non-zero elements.
In the backward pass, these activations are decompressed in order to calculate the gradients with respect to the activations as part of the backpropagation algorithm.
The bitmap format represents the minimum perceivable memory required to store the information in a matrix; that is non-zero elements (represented as half-, single- or double-precision) and a single bit for each element index.
We denote the memory footprint as $M_d$ and $M_b$ for stashing activations in a dense format and bitmap format respectively.
For single-precision (FP32), the memory footprint (in bytes) would be calculated as:

\begin{center}
    $M_d$ = $4 \times$ total activations \\
    $M_b$ = $4 \times$ non-zero activations + (1/8) $\times$ total activations    
\end{center}

We also compare the bitmap format with the COO format, which represents indices as either integers (4 bytes) or longs (8 bytes).
We denote the memory footprint (in bytes) of the COO format as $M_c$, and it can be calculated as:

\begin{algorithm}[t!]
    \small
    \SetKwBlock{Begin}{begin}{end}
    \SetKwInOut{Input}{Input}
    \SetKwInOut{Output}{Output}
    \Input{Dense Activation Matrix ($T$)}
    \Output{Bitmap Matrix ($B$)}
    $B$.shape = $T$.shape  // deep copy \\
    Flatten $T$ \\
    \For{$i \; in \; 0, .., length(T)$}{
        \eIf{$T[i] == 0$}{
            Push $0$ bit to $B$.bitmap
        }
        {
            Push $T[i]$ to $B$.values \\
            Push $1$ bit to $B$.bitmap
        }
    }
    delete $T$  // free dense memory
    \caption{Dense to Bitmap Matrix Compression}
    \label{alg:tobitmap}
\end{algorithm}

\begin{algorithm}[t!]
    \small
    \SetKwBlock{Begin}{begin}{end}
    \SetKwInOut{Input}{Input}
    \SetKwInOut{Output}{Output}
    \Input{Bitmap Matrix ($B$)}
    \Output{Dense Activation Matrix ($T$)}
    Construct 1-dimensional $T$; initialize $j = 0$ \\
    \For{$i \; in \; 0, .., length(B)$.bitmap}{
        \eIf{$B$.bitmap[i] is set}{
            Push $B$.values[j] to $T$; increment j \\
        }
        {
            Push $0.0$ to $T$  // half-, single- or double-precision
        }
    }
    reshape $T$ to $B$.shape \\
    delete $B$  // free bitmap memory
    \caption{Bitmap to Dense Matrix Decompression}
    \label{alg:todense}
\end{algorithm}

\begin{figure}[t!]
    \centering
    \includegraphics[width=0.5\textwidth]{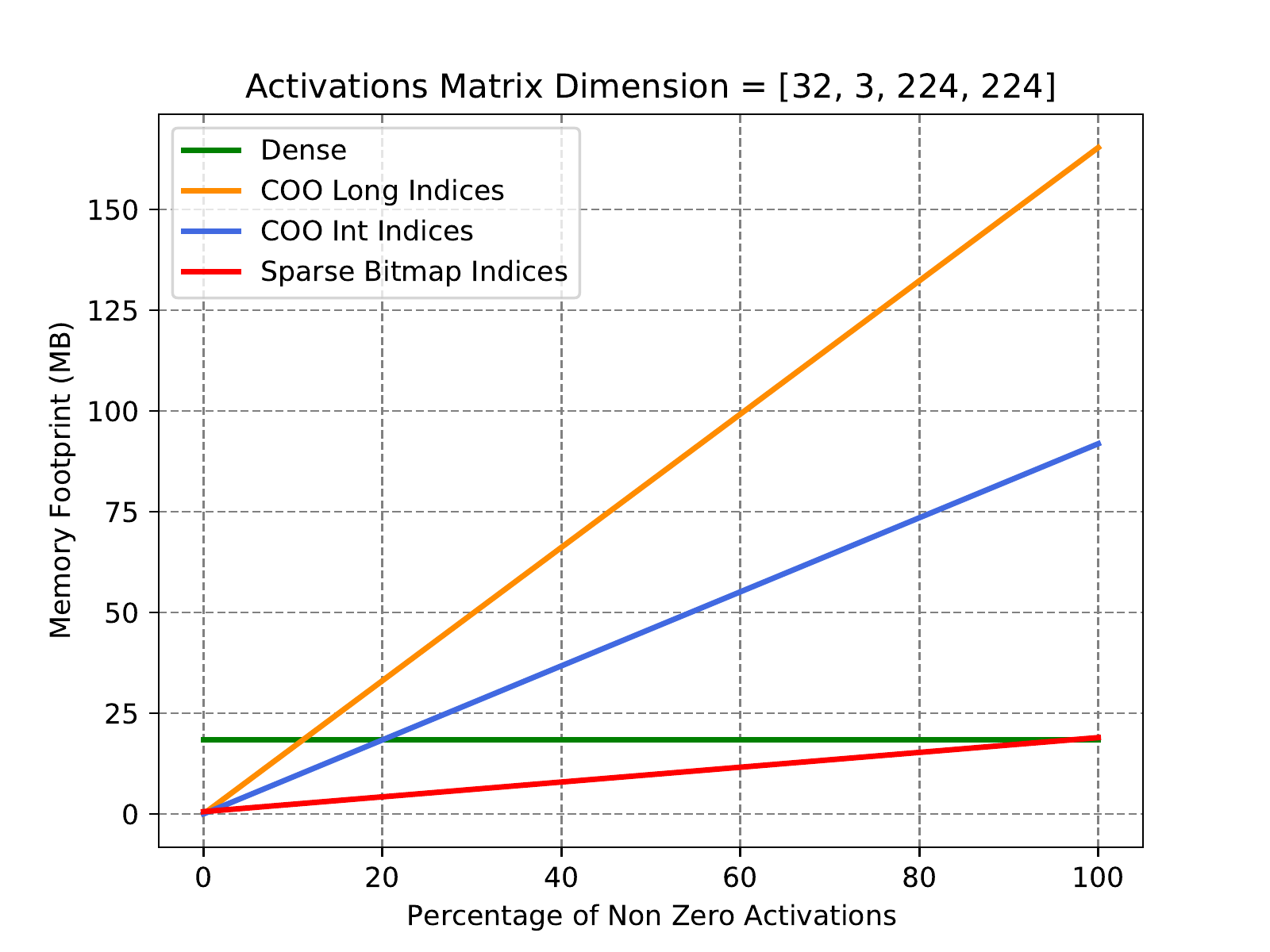}
    \belowcaptionskip -10 pt
    \caption{Memory footprint of our proposed bitmap format as compared to other sparse matrices.}
    \label{fig:sparse-matrices-memory-footprint}
\end{figure}

\begin{center}
    $M_c$ = $(4 + [4|8] \times$ num-dimensions $) \times$ non-zero activations
\end{center}
\noindent where $4$ is the size of single-precision for saving the activation values, and $[4|8]$ are the sizes for either integer or long indices.
For example, PyTorch and Tensorflow use long indices by default in their COO implementations.
Figure \ref{fig:sparse-matrices-memory-footprint} shows that the COO format is only efficient if the activations have a sparsity of at least 80\% (~20\% non-zero activations). 
We observe that the dense representation consistently maintains a low memory footprint.
However, the proposed sparse bitmap format can reduce the memory footprint even if the activations matrix has low sparsity.
Unlike other sparse matrix formats, the sparse bitmap format is used for stashing the activations until they are needed in the backward pass, and not for directly operating on them (e.g. multiplication).

\subsection{Sparse Bitmap Compression Algorithm}
In order to achieve empirical memory reductions, our algorithm avoids copying matrices in function calls.
Compressing a dense matrix to a bitmap matrix is outlined in Algorithm \ref{alg:tobitmap}.
We first start by keeping a copy of the shape (dimension sizes) of the original matrix.
Operating on a vector is also necessary to avoid complex index resolution operations, so we flatten the dense matrix.
This is an in-place operation that does not move values to different memory locations.
Lines 3 to 10 scans the vector, constructing the bitmap and stashing the non-zero elements.
Finally, line 11 frees the memory used by the dense matrix to be used in the subsequent layers.
Algorithm \ref{alg:todense} performs the opposite operation in the backward pass.
In particular, it maps the non-zero elements to a dense vector matching the bitmap, and then reshapes the data (in-place) according to the previously saved shape.

Note that both algorithms are linear runtime.
They take $O(n)$ time, where $n$ is the total number of activations in a matrix.
In the next section, we provide additional implementation details for a modern deep learning framework.

\section{Experiments}
\label{sec_experiments}
Our experimental analysis tests the hypothesis of memory reduction using the sparse bitmap format both theoretically and empirically.
First, we analyze the memory footprint reduction in state-of-the-art CNN-based architectures using both ImageNet \cite{deng2009imagenet} and CIFAR 10 \cite{krizhevsky2014cifar} image datasets.
Afterwards, we study the compound reduction in training memory footprint when combining BitTrain with other orthogonal training methods such as low precision training, activation pruning, and checkpointing.
Finally, we present our own implementation of BitTrain, and analyze the on-board memory footprint reduction as well as the runtime overhead.

\textbf{Setup. }
In our experiments, we use PyTorch version 1.7, and in the C++ implementation, we use libtorch version 1.7.
We use Clang version 10.0.0 as the compiler, and compile using C++14 standards.
Empirical memory footprint measurements are performed on Nvidia Jetson Nano board that has 4GB of memory.
More details on the implementation and memory measurements is provided below.

\begin{figure}[t]
\centering
\begin{tabular}{c}
\subfloat[\label{class_models_imagenet}]{\includegraphics[width=0.45\textwidth, keepaspectratio]{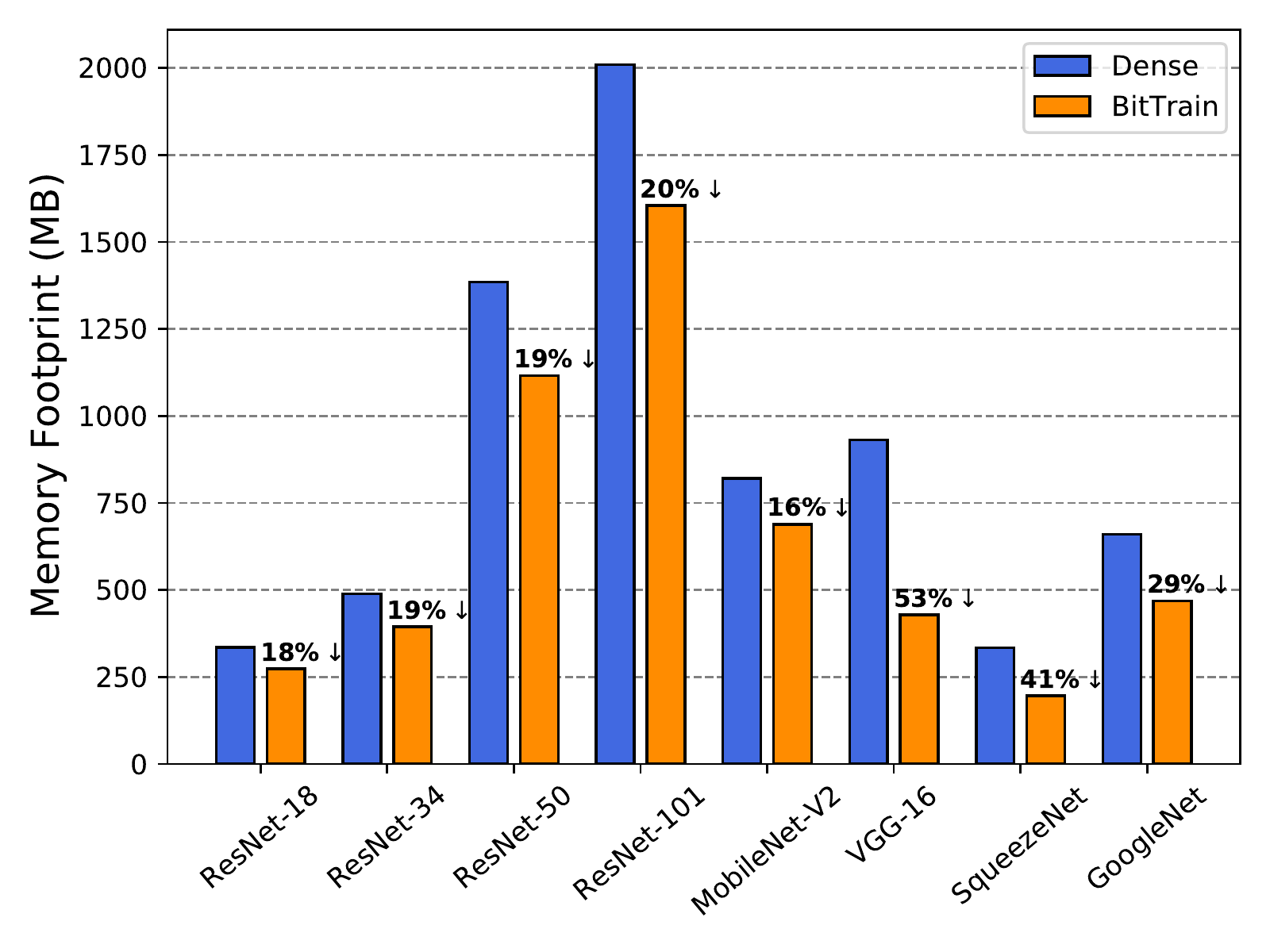}} \\
\subfloat[\label{class_models_cifar10}]{\includegraphics[width=0.45\textwidth, keepaspectratio]{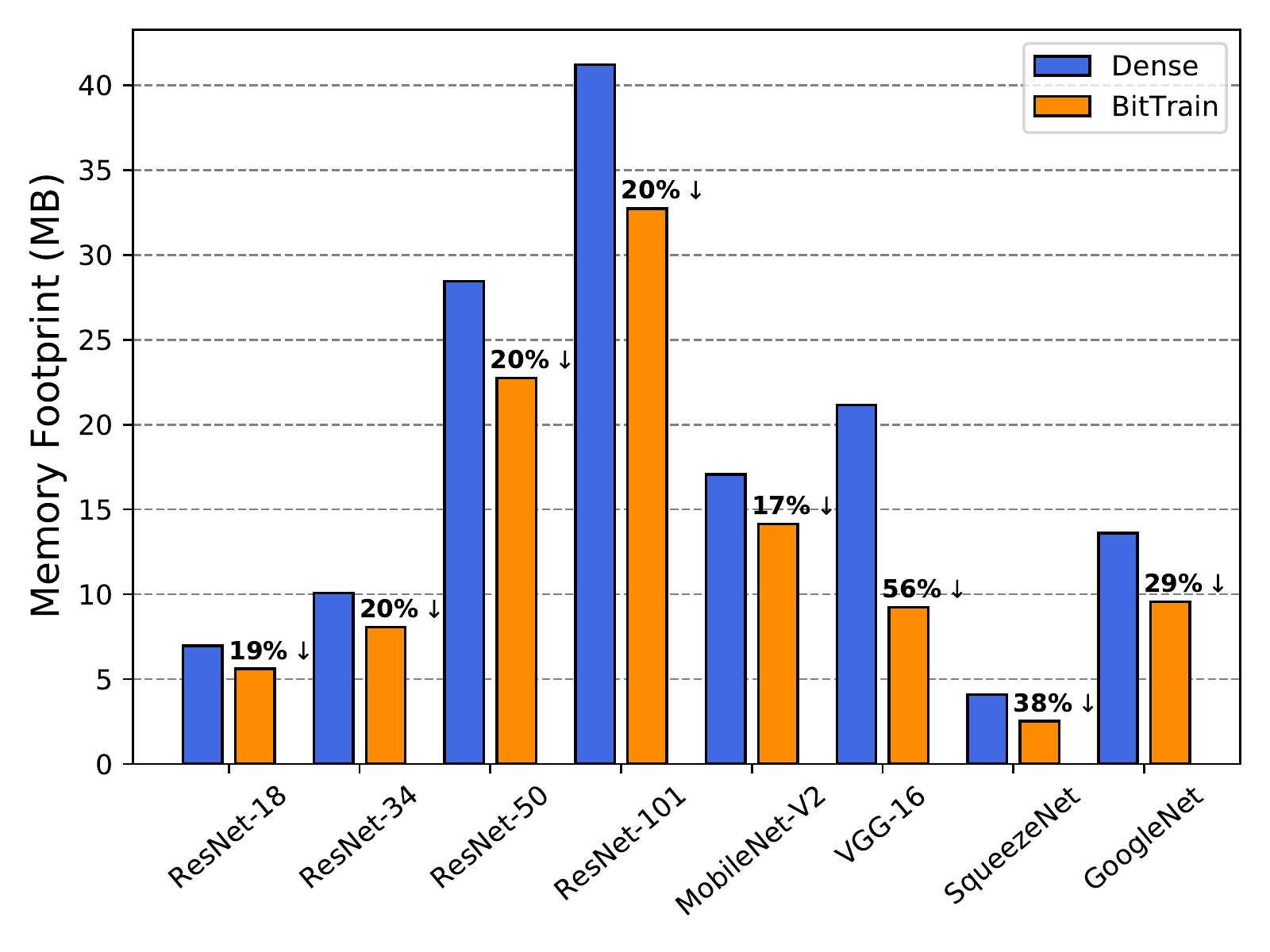}}
\end{tabular}
\belowcaptionskip -10 pt
\caption{Activations Memory footprint for training different classification models on (a) ImageNet (b) Cifar10.}
\label{fig:memory-foot-print-reduction}
\end{figure}

\textbf{Classification Models.}
Our sparse bitmap format reduces the memory footprint for storing activations with high sparsity as previously illustrated in Figure \ref{fig:sparse-matrices-memory-footprint}.
To access the overall impact of using our proposed methodology on the memory footprint during training, we choose 8 state-of-the-art classification models that vary in size and complexity. 
We analyze the training memory footprint of those models for two different classification datasets: CIFAR-10 (Input resolution is $32 \times 32$), and ImageNet (Input resolution is $224 \times 224$). 
Figure \ref{fig:memory-foot-print-reduction} shows the memory footprint of BitTrain in comparison to classical training (referred to as \textit{Dense}). Figures \ref{class_models_imagenet} and \ref{class_models_cifar10} represent training different classification models on ImageNet, and CIFAR-10 respectively.
The results show that BitTrain reduces the memory footprint by up to 56\%, this improvement is achieved by leveraging the activations sparsity that are naturally found in those models.
We can notice that the improvements tends to be higher for VGG-16, SqueezeNet, and GoogleNet.
The reason is that those models tend to have higher sparsity percentages because they do not have any batch normalization layers.
In the following Section, we propose using activations pruning to increase the activations sparsity, and hence maximize the memory footprint reduction that could be achieved by our sparse bitmap compression technique.

\begin{figure}[t!]
    \centering
    \includegraphics[width=0.45\textwidth]{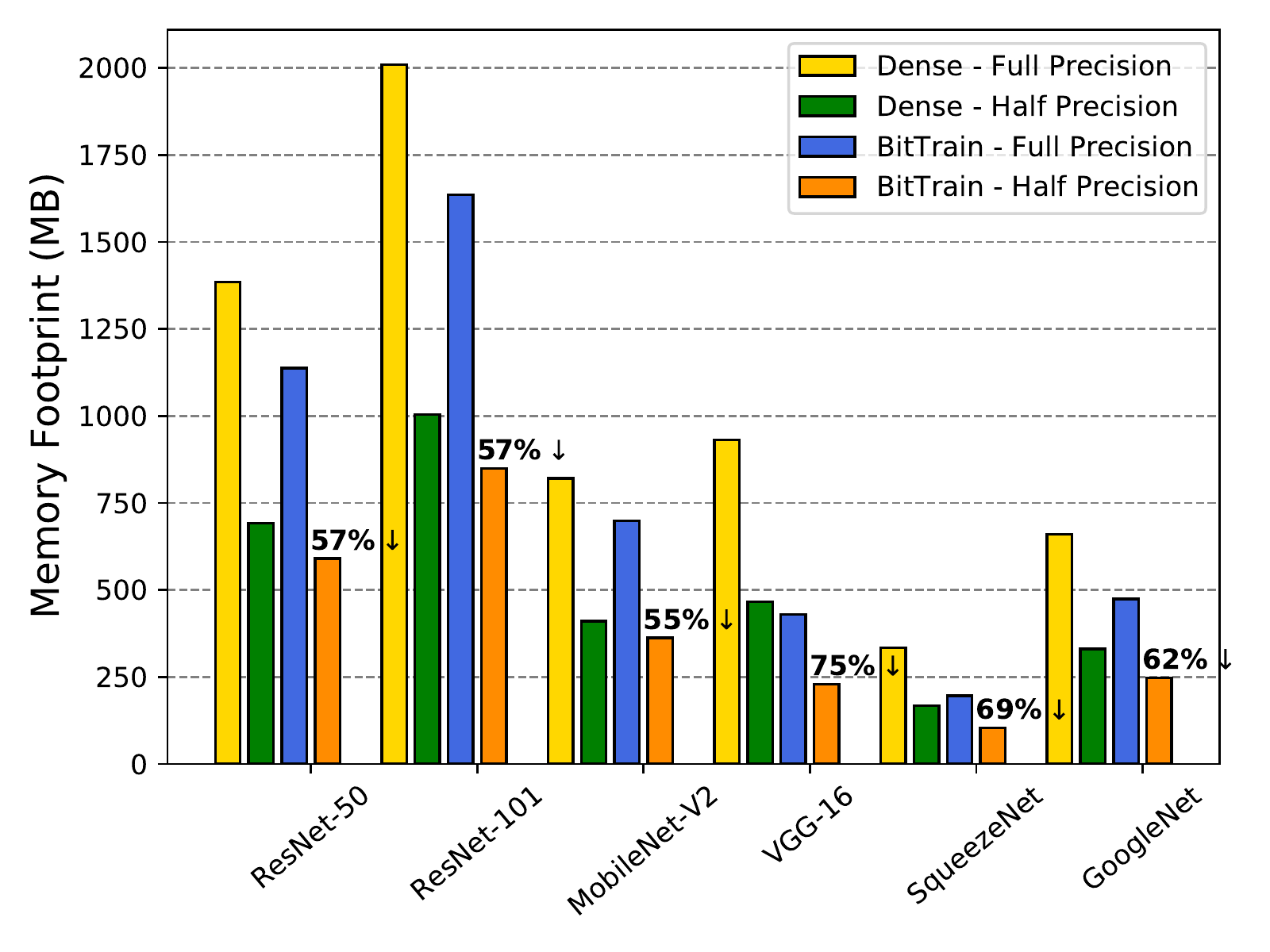}
    \caption{Activations Memory footprint for training on ImageNet when using half-precision (FP16) for storing the activations.}
    \label{fig:half-precision}
\end{figure}

% Also describe reduction in accuracy if any. 
\textbf{Combining BitTrain with Low Precision.}
Neural network parameters are typically represented in single-precision (FP32) [IEEE 32-bit].
Using half-precision (FP16) [IEEE 16-bit] has shown to be sufficient for the general case of training neural network as discussed in Section \ref{sec_relatedwork}.
It reduces the memory footprint of all components (model parameters, activations, and optimizer gradients).
Low precision training is supported in modern deep learning frameworks, and is as straightforward as specifying the $float16$ data type for all parameters and model inputs.
Figure \ref{fig:half-precision} shows the memory footprint reduction that can be achieved when combining our bitmap format for saving the activations with half-precision arithmetic.
We observe that combining using half-precision activation values with our sparse bitmap compression offers a 55-75\% saving in the memory required for storing the activations when compared to the dense full-precision baseline. 
This means that using our proposed technique achieves up to an additional 25\% when used with half-precision than using half-precision alone. 
As for the accuracy, Sohoni et al. \cite{lowmemory} show that training with half-precision does not meaningfully affect the final accuracy.
It is important to note that half-precision requires native hardware support in order to achieve empirical results \citep{durant-half-precision}.

\begin{figure*}[t]
\centering
\begin{tabular}{c}
\subfloat[\label{activ_pruning_memory_imagenet}]{\includegraphics[width=0.5\textwidth, keepaspectratio]{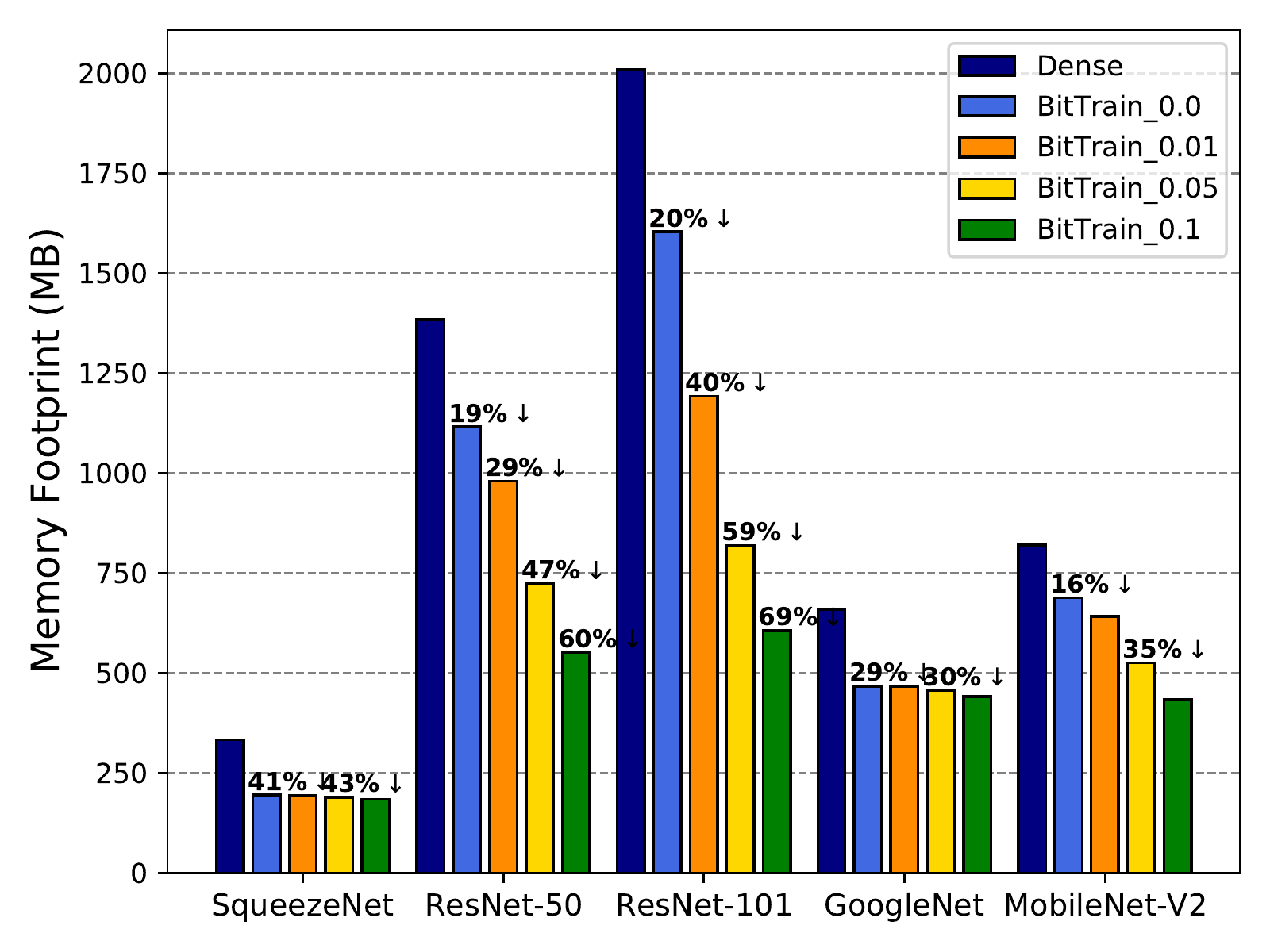}}
\subfloat[\label{activ_pruning_memory_cifar10}]{\includegraphics[width=0.5\textwidth, keepaspectratio]{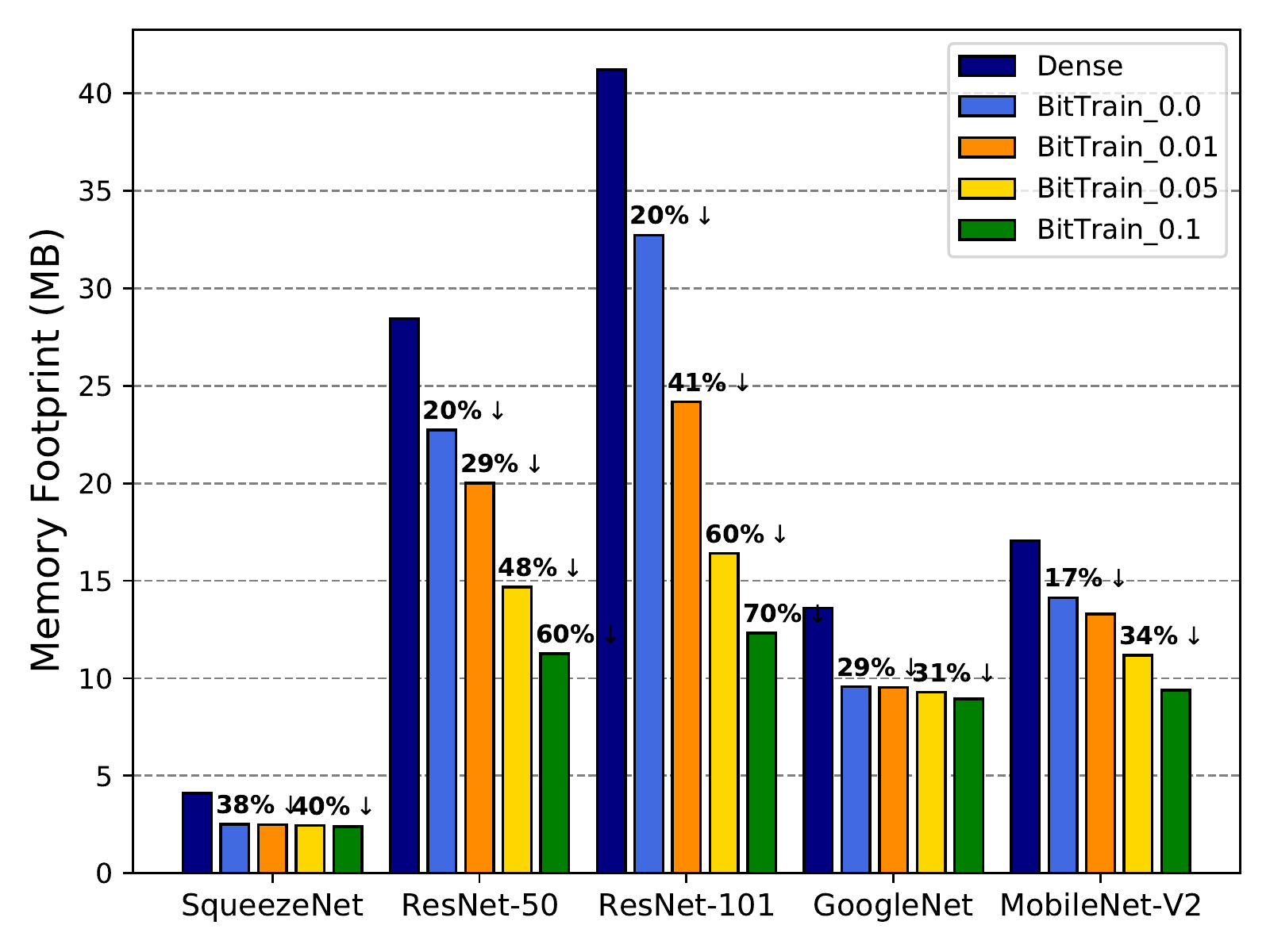}}
\end{tabular}
\belowcaptionskip -10 pt
\caption{Activations pruning analysis for classification models with different pruning thresholds. In all figures $Bitmap\_x$ denotes that our sparse bitmap compression is used along with activation pruning with threshold $x$ (a) Training Memory footprint for ImageNet (b) Training Memory footprint for CIFAR-10.}
\label{fig:memory-foot-print-reduction-activation-pruning}
\end{figure*}

\textbf{Combining BitTrain with Activation Pruning.}
As illustrated in Figure \ref{fig:activations-histogram}, more than 70\% of the activations have values that are close to zero.
This means that 70\% of the activations are not effective during the training process.
We leverage this fact to further increase the activations sparsity which would further improve the memory footprint when using our sparse bitmap compression.
In our implementation, we only store the activations that exceed a certain pre-defined close-to-zero threshold. 
However, if the activation value is less than the threshold, we prune this value (i.e. set it to zero). 
In Figure \ref{fig:memory-foot-print-reduction-activation-pruning}, we analyze the effect of using activation pruning along with BitTrain on the training memory footprint.
Figures \ref{activ_pruning_memory_imagenet} and \ref{activ_pruning_memory_cifar10} show the training memory footprint for five different models on ImageNet, and CIFAR-10 respectively. 
We analyze the memory savings using different activation pruning thresholds ($0$, $0.01$, $0.05$, $0.1$). 
The results shows that memory footprint reduction increases as the activation pruning threshold increases.
This is expected because increasing the activation pruning threshold increases the percentage on zero elements in the model, which maximizes the gains from using our sparse bitmap compression technique.
We can notice that the gains from using activation pruning varies from one model to another depending on the percentage of close-to-zero activations in the model.
For example, using activation pruning with ResNet-50 and ResNet-101 achieves up to an additional 49\% reduction in memory footprint, while it provide insignificant memory footprint reduction when applied to GoogleNet.

\begin{figure}[t!]
    \centering
    \includegraphics[width=0.5\textwidth]{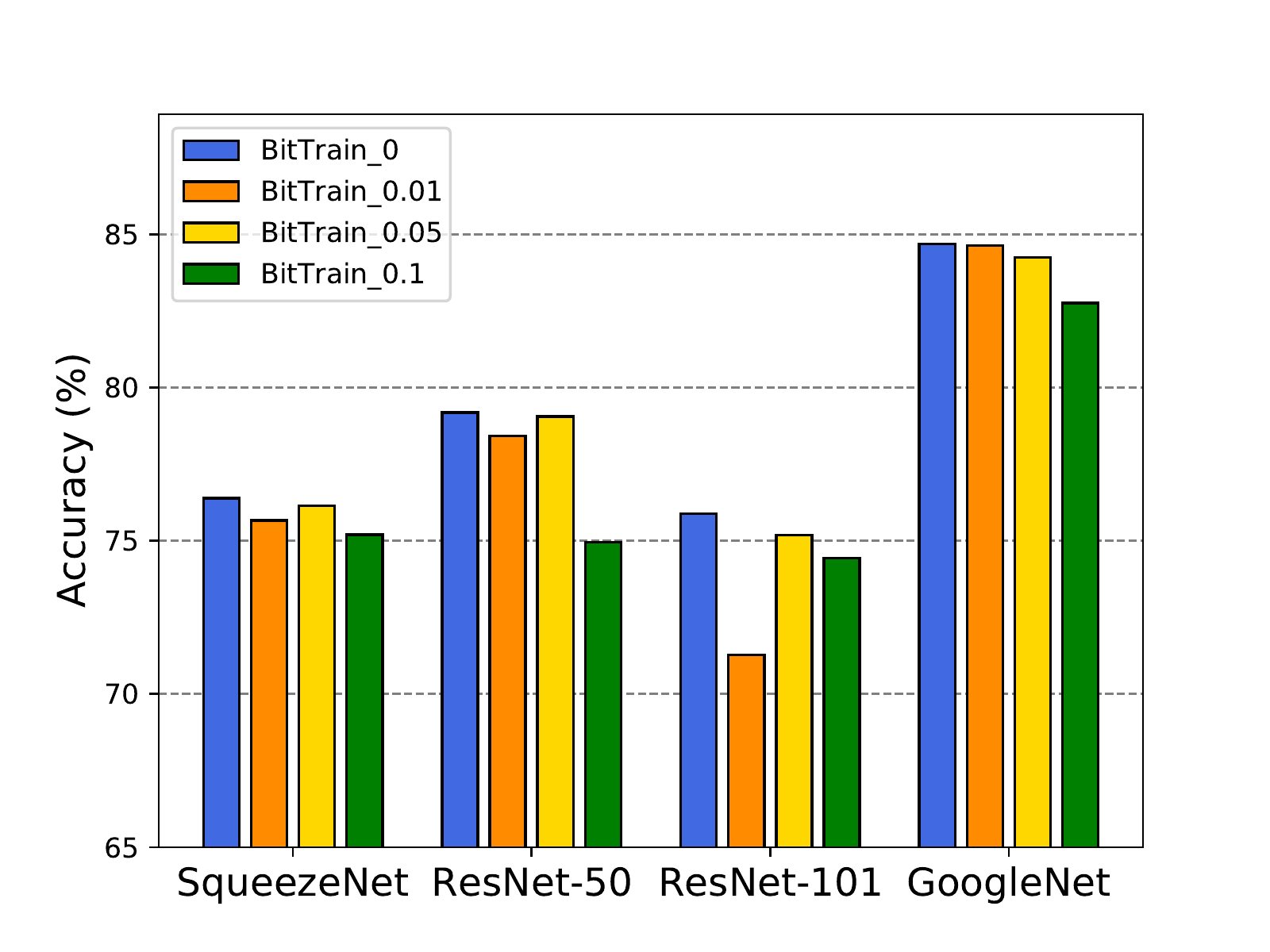}
    \belowcaptionskip -15 pt
    \caption{Classification Accuracy on Cifar-10 for different models when training under different activation pruning thresholds.}
    \label{fig:accuracy_activ_pruning_cifar10}
\end{figure}

We also analyze the accuracy of using BitTrain along with the activation pruning on CIFAR-10 in Figure \ref{fig:accuracy_activ_pruning_cifar10}.
We can see that the accuracy drop varies between different models. 
The accuracy drop depends on the significance of the pruned values, and how the model training adapts to the pruning. 
This creates a memory-accuracy trade-off. 
For some models like ResNet-50, it might be worth it to trade a negligible loss in accuracy for up to a 49\% reduction in the memory footprint. 
However, it might not be worth it for models like GoogleNet, where using the activation pruning achieves a modest reduction in the memory footprint.

\begin{figure}[t!]
    \centering
    \includegraphics[width=0.5\textwidth]{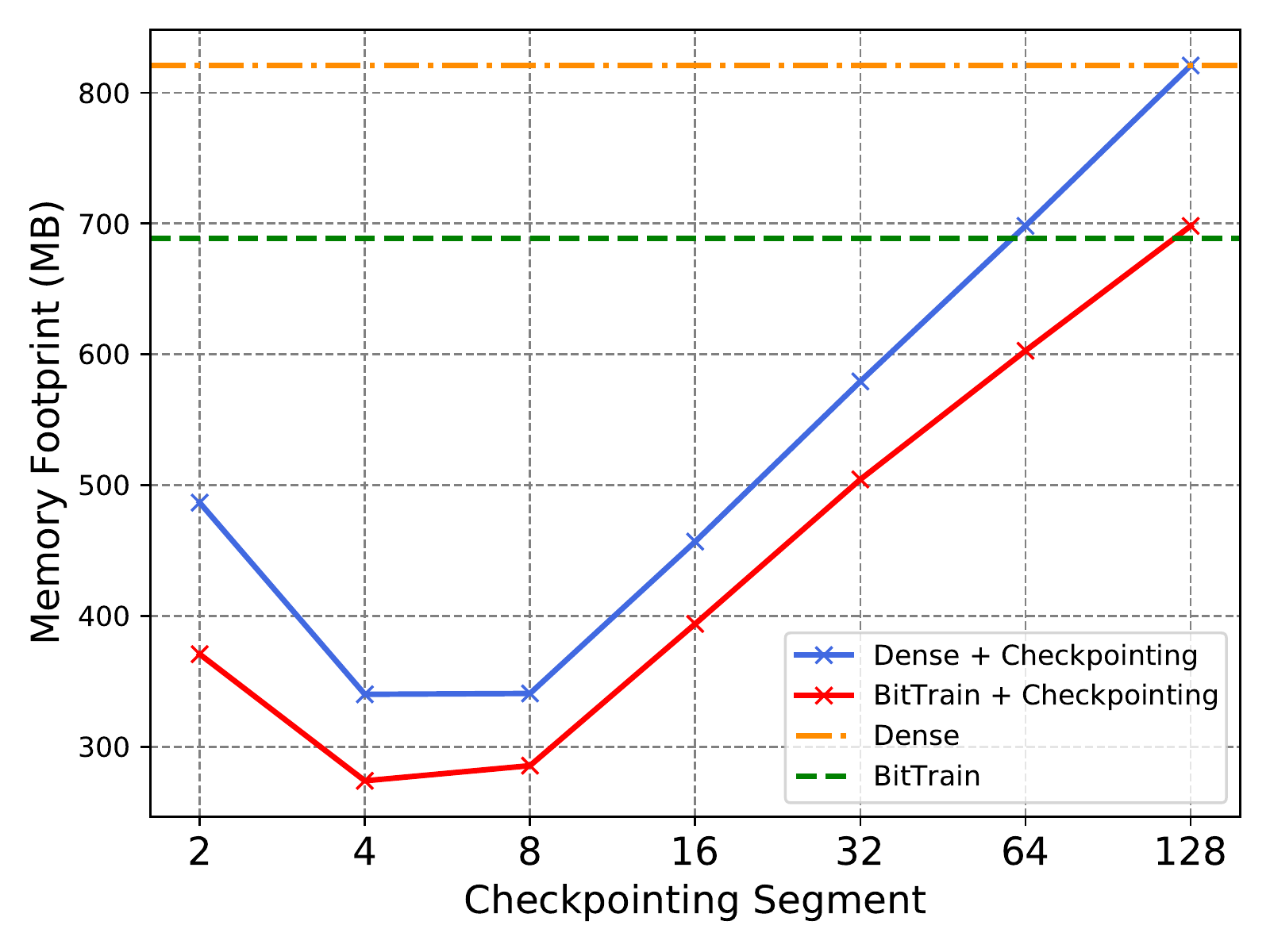}
    \caption{Memory footprint reduction for the activations when using checkpointing for partial storage of the activations for MobileNet-V2 when trained on ImageNet.}
    \label{fig:with_checkpointing}
\end{figure}

\textbf{Combining BitTrain with Checkpointing.}
Checkpointing is used in literature to trade computations for memory. 
The idea is to only store some of the intermediate activations, and re-compute the others during backpropagation. 
In this section, we implement the checkpointing algorithm, then combine it with BitTrain to analyze the compound memory savings. 
We implement the \textit{checkpoint-every-m} checkpointing strategy as it is the commonly-used approach for checkpointing \citep{lowmemory}.
In the \textit{checkpoint-every-m} strategy, the input activations of every $m$ layers are stored during the forward pass.
For example, assume that we have a simple feedforward model with $m \times n$ layers. 
During the forward pass, we store the activations for one layer every $m$ layers.
This divides the model into \textit{n} segments, where each segment has one layer with stored input activation (i.e., we store the input activations for n layers).
During the backward pass, we recompute the activations of all the layers for each segment, and we store them temporarily in order to compute gradients with respect to the layers within the segment. 
After this, we discard the temporarily stored activations and proceed to the next segment.
We combine the \textit{checkpoint-every-m} checkpointing strategy with BitTrain, and analyze the compound effect on the memory footprint as shown in Figure \ref{fig:with_checkpointing}. 
Using our sparse bitmap compression, we can achieve up to an extra 25\% reduction in memory footprint compared to checkpointing alone.

\begin{table*}[t!]
\caption{Memory footprint reduction when using our proposed bitmap format for storing convolution activations. Convolution sizes are chosen as they appear in order in the ResNet model. On-board is executed a Jetson Nano board.}

\label{tab:memory-footprint}
\small
\renewcommand{\arraystretch}{1.05}
\setlength{\tabcolsep}{1.5pt}
\begin{tabular}{@{}lllllrrrrrrr@{}}
\toprule
\multicolumn{1}{c}{\multirow{2}{*}{\textbf{\begin{tabular}[c]{@{}c@{}}Batch\\ Size\end{tabular}}}} & \multicolumn{1}{c}{\multirow{2}{*}{\textbf{Channels}}} & \multicolumn{1}{c}{\multirow{2}{*}{\textbf{Width}}} & \multicolumn{1}{c}{\multirow{2}{*}{\textbf{Height}}} & \multicolumn{1}{c}{\multirow{2}{*}{\textbf{\begin{tabular}[c]{@{}c@{}}num \\ elements\end{tabular}}}} & \multicolumn{1}{c}{\multirow{2}{*}{\textbf{\begin{tabular}[c]{@{}c@{}}\%non-\\ zeros\end{tabular}}}} & \multicolumn{2}{c}{\textbf{Dense Tensor (MB)}} & \multicolumn{4}{c}{\textbf{Bitmap Tensor (MB)}} \\ \cmidrule(l){7-12} 
\multicolumn{1}{c}{} & \multicolumn{1}{c}{} & \multicolumn{1}{c}{} & \multicolumn{1}{c}{} & \multicolumn{1}{c}{} & \multicolumn{1}{c}{} & \multicolumn{1}{c}{\textbf{Theoretical}} & \multicolumn{1}{c}{\textbf{On-board}} & \multicolumn{1}{c}{\textbf{Theoretical}} & \multicolumn{1}{c}{\textbf{Improv (\%)}} & \multicolumn{1}{c}{\textbf{On-board}} & \multicolumn{1}{c}{\textbf{Improv (\%)}} \\ \midrule
\multicolumn{1}{r}{16} & \multicolumn{1}{r}{3} & \multicolumn{1}{r}{224} & \multicolumn{1}{r}{224} & \multicolumn{1}{r}{2,408,448} & \multicolumn{1}{r|}{0\%} & 9.19 & \multicolumn{1}{r|}{10.53} & 0.29 & \multicolumn{1}{r|}{96.88} & 2.50 & 76.26 \\
 &  &  &  &  & \multicolumn{1}{r|}{25\%} & 9.19 & \multicolumn{1}{r|}{10.77} & 2.58 & \multicolumn{1}{r|}{71.88} & 5.14 & 52.27 \\
 &  &  &  &  & \multicolumn{1}{r|}{50\%} & 9.19 & \multicolumn{1}{r|}{10.83} & 4.88 & \multicolumn{1}{r|}{46.88} & 7.47 & 31.02 \\
 &  &  &  &  & \multicolumn{1}{r|}{75\%} & 9.19 & \multicolumn{1}{r|}{10.87} & 7.18 & \multicolumn{1}{r|}{21.88} & 9.71 & 10.67 \\
 &  &  &  &  & \multicolumn{1}{r|}{100\%} & 9.19 & \multicolumn{1}{r|}{10.76} & 9.47 & \multicolumn{1}{r|}{-3.13} & 12.03 & -11.80 \\ \midrule
\multicolumn{1}{r}{16} & \multicolumn{1}{r}{7} & \multicolumn{1}{r}{112} & \multicolumn{1}{r}{112} & \multicolumn{1}{r}{1,404,928} & \multicolumn{1}{r|}{0\%} & 5.36 & \multicolumn{1}{r|}{7.17} & 0.17 & \multicolumn{1}{r|}{96.88} & 2.67 & 62.76 \\
 &  &  &  &  & \multicolumn{1}{r|}{25\%} & 5.36 & \multicolumn{1}{r|}{7.11} & 1.51 & \multicolumn{1}{r|}{71.88} & 4.07 & 42.76 \\
 &  &  &  &  & \multicolumn{1}{r|}{50\%} & 5.36 & \multicolumn{1}{r|}{6.96} & 2.85 & \multicolumn{1}{r|}{46.88} & 5.19 & 25.43 \\
 &  &  &  &  & \multicolumn{1}{r|}{75\%} & 5.36 & \multicolumn{1}{r|}{7.4} & 4.19 & \multicolumn{1}{r|}{21.88} & 6.97 & 5.81 \\
 &  &  &  &  & \multicolumn{1}{r|}{100\%} & 5.36 & \multicolumn{1}{r|}{7.17} & 5.53 & \multicolumn{1}{r|}{-3.13} & 8.11 & -13.11 \\ \midrule
\multicolumn{1}{r}{16} & \multicolumn{1}{r}{64} & \multicolumn{1}{r}{56} & \multicolumn{1}{r}{56} & \multicolumn{1}{r}{3,211,264} & \multicolumn{1}{r|}{0\%} & 12.25 & \multicolumn{1}{r|}{13.95} & 0.38 & \multicolumn{1}{r|}{96.88} & 3.03 & 78.28 \\
 &  &  &  &  & \multicolumn{1}{r|}{25\%} & 12.25 & \multicolumn{1}{r|}{13.95} & 3.45 & \multicolumn{1}{r|}{71.88} & 6.1 & 56.27 \\
 &  &  &  &  & \multicolumn{1}{r|}{50\%} & 12.25 & \multicolumn{1}{r|}{13.86} & 6.51 & \multicolumn{1}{r|}{46.88} & 9.12 & 34.20 \\
 &  &  &  &  & \multicolumn{1}{r|}{75\%} & 12.25 & \multicolumn{1}{r|}{13.48} & 9.57 & \multicolumn{1}{r|}{21.88} & 11.77 & 12.69 \\
 &  &  &  &  & \multicolumn{1}{r|}{100\%} & 12.25 & \multicolumn{1}{r|}{13.85} & 12.63 & \multicolumn{1}{r|}{-3.13} & 15.24 & -10.04 \\ \midrule
\multicolumn{1}{r}{16} & \multicolumn{1}{r}{128} & \multicolumn{1}{r}{28} & \multicolumn{1}{r}{28} & \multicolumn{1}{r}{1,605,632} & \multicolumn{1}{r|}{0\%} & 6.13 & \multicolumn{1}{r|}{7.41} & 0.19 & \multicolumn{1}{r|}{96.88} & 2.10 & 71.66 \\
 &  &  &  &  & \multicolumn{1}{r|}{25\%} & 6.13 & \multicolumn{1}{r|}{7.64} & 1.72 & \multicolumn{1}{r|}{71.88} & 4.02 & 47.38 \\
 &  &  &  &  & \multicolumn{1}{r|}{50\%} & 6.13 & \multicolumn{1}{r|}{7.38} & 3.25 & \multicolumn{1}{r|}{46.88} & 5.31 & 28.05 \\
 &  &  &  &  & \multicolumn{1}{r|}{75\%} & 6.13 & \multicolumn{1}{r|}{7.65} & 4.79 & \multicolumn{1}{r|}{21.88} & 6.91 & 9.67 \\
 &  &  &  &  & \multicolumn{1}{r|}{100\%} & 6.13 & \multicolumn{1}{r|}{7.92} & 6.32 & \multicolumn{1}{r|}{-3.13} & 8.83 & -11.49 \\ \midrule
\multicolumn{1}{r}{16} & \multicolumn{1}{r}{256} & \multicolumn{1}{r}{14} & \multicolumn{1}{r}{14} & \multicolumn{1}{r}{802,816} & \multicolumn{1}{r|}{0\%} & 3.06 & \multicolumn{1}{r|}{4.80} & 0.10 & \multicolumn{1}{r|}{96.88} & 2.36 & 50.83 \\
 &  &  &  &  & \multicolumn{1}{r|}{25\%} & 3.06 & \multicolumn{1}{r|}{4.61} & 0.86 & \multicolumn{1}{r|}{71.88} & 3.16 & 31.45 \\
 &  &  &  &  & \multicolumn{1}{r|}{50\%} & 3.06 & \multicolumn{1}{r|}{4.99} & 1.63 & \multicolumn{1}{r|}{46.88} & 4.39 & 12.02 \\
 &  &  &  &  & \multicolumn{1}{r|}{75\%} & 3.06 & \multicolumn{1}{r|}{4.95} & 2.39 & \multicolumn{1}{r|}{21.88} & 5.11 & -3.23 \\
 &  &  &  &  & \multicolumn{1}{r|}{100\%} & 3.06 & \multicolumn{1}{r|}{4.33} & 3.16 & \multicolumn{1}{r|}{-3.13} & 5.24 & -21.02 \\ \midrule
\multicolumn{1}{r}{16} & \multicolumn{1}{r}{512} & \multicolumn{1}{r}{7} & \multicolumn{1}{r}{7} & \multicolumn{1}{r}{401,408} & \multicolumn{1}{r|}{0\%} & 1.53 & \multicolumn{1}{r|}{3.55} & 0.05 & \multicolumn{1}{r|}{96.88} & 2.47 & 30.42 \\
 &  &  &  &  & \multicolumn{1}{r|}{25\%} & 1.53 & \multicolumn{1}{r|}{3.26} & 0.43 & \multicolumn{1}{r|}{71.88} & 3.02 & 7.36 \\
 &  &  &  &  & \multicolumn{1}{r|}{50\%} & 1.53 & \multicolumn{1}{r|}{3.52} & 0.81 & \multicolumn{1}{r|}{46.88} & 3.55 & -0.85 \\
 &  &  &  &  & \multicolumn{1}{r|}{75\%} & 1.53 & \multicolumn{1}{r|}{3.51} & 1.20 & \multicolumn{1}{r|}{21.88} & 4.08 & -16.24 \\
\multicolumn{1}{r}{} & \multicolumn{1}{r}{} & \multicolumn{1}{r}{} & \multicolumn{1}{r}{} & \multicolumn{1}{r}{} & 100\% & 1.53 & 3.24 & 1.58 & -3.13 & 4.16 & -28.40 \\ \bottomrule
\end{tabular}
\end{table*}

\textbf{Implementation Details.}
High-level languages used in the deep learning frameworks do not provide fine-grained memory management APIs.
For example, Python depends on garbage collection techniques the frees up memory of a given object (i.e. tensor or matrix) when there is no references to it \cite{van1995python}.
This leaves very little room to the developer in controlling how tensors are stored in memory.
Moreover, all data types in Python are of type \textit{PyObject}, which means that numbers, characters, strings, and bytes are actually Python objects that consumes more memory for object metadata in order to be tracked by the garbage collector.
In other words, defining bits or bytes and expecting to get accurate memory measurements is infeasible.
Therefore, we implemented BitTrain in C++, using $bitset$ and $vector$ data types from the C++ standard library for storing the bitmap and the non-zero activations respectively.
Our implementation extends libtorch's C++ API \cite{paszke2019pytorch}, by defining a tensor that inherits from the default tensor implementation.
We chose libtorch because its tensor definition separates the tensor storage from its definition in the computation graph, allowing us to implement Algorithms \ref{alg:tobitmap} and \ref{alg:todense}.
Furthermore, it allows our tensor to integrate natively to its dynamic computation graph.

\textbf{Measuring Memory Footprint.}
Advances in memory hierarchies (i.e., cache, memory, virtual memory) has made it challenging to measure the exact memory consumed by training a neural network.
In addition, deep learning frameworks heavily depend on shared libraries on the host system that can be used by other processes.
In BitTrain, we measure the memory footprint using the Unique Set Size (USS) of the running process. 
USS is the memory that is unique to the process and which would be freed if the process was to be terminated at the moment of measurement. 
On Linux, we calculate this value by parsing all the private blocks in $/proc/pid/smaps$.
We note that previous methods described in \citep{tinytl, dynamic, sparsetraining} do not provide implementations, and do not measure the actual memory footprint.
Rather, they only present approximate calculations from the PyTorch APIs.
Since BitTrain is mainly focusing on edge devices, we show how the implementation is compared to the theoretical estimations.

% x-axis: number of elements
% y-axis: memory
% each line is a separate sparsity level

\textbf{Activations Memory Reduction.}
Table \ref{tab:memory-footprint} shows the memory reduction per convolution activations as compared to the calculated results.
Although a batch size of 32 is more stable for training \citep{smith2017don}, we chose a batch size of 16 as a more realistic benchmark for training on the edge.
We tested our compression against convolutional layer sizes (number of channels, width and height of activation maps) in ResNet, which can be representative of many convolution layer sizes in the literature.
First, we observe that while the empirical gain deviates from the theoretical calculations, the implementation is still efficient at different sparsity levels.
For example, at 50\% sparsity, our method achieves up to 34\% memory reduction.
According to Figure \ref{fig:activations-histogram} in Section \ref{sec_methodology}, sparsity can be expected to be more than 70\%.
In this case, we save activations memory by up to 56\% depending on the size of the activations.

\begin{figure}[t!]
    \centering
    \includegraphics[width=0.5\textwidth]{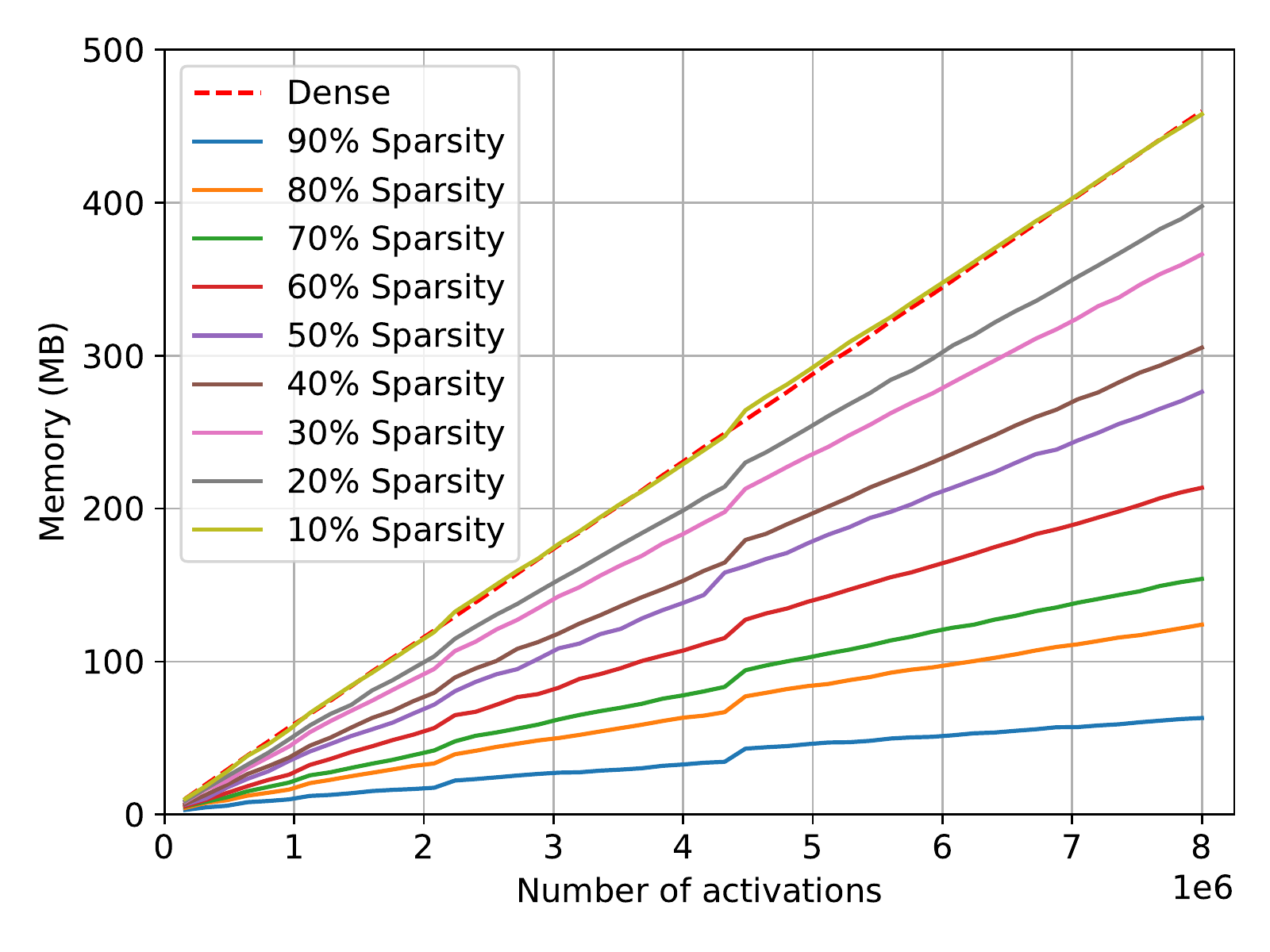}

    \caption{On-board (Jetson Nano) memory reduction as a function of the activation size and sparsity level (implementation using libtorch C++ API).}

    \label{fig:implementation-memory-reduction}
\end{figure}

Moreover, we analyzed how the bitmap format scales with the increasing number of activations.
Figure \ref{fig:implementation-memory-reduction} shows that it scales sub-linearly as compared to the saving activations in a dense format.
We observe that memory savings is proportional to the number of activations and the sparsity level.
This proves that BitTrain is a step towards to enabling training modern convolutional neural networks on edge devices.

\begin{figure}[t!]
    \centering
    \includegraphics[width=0.5\textwidth]{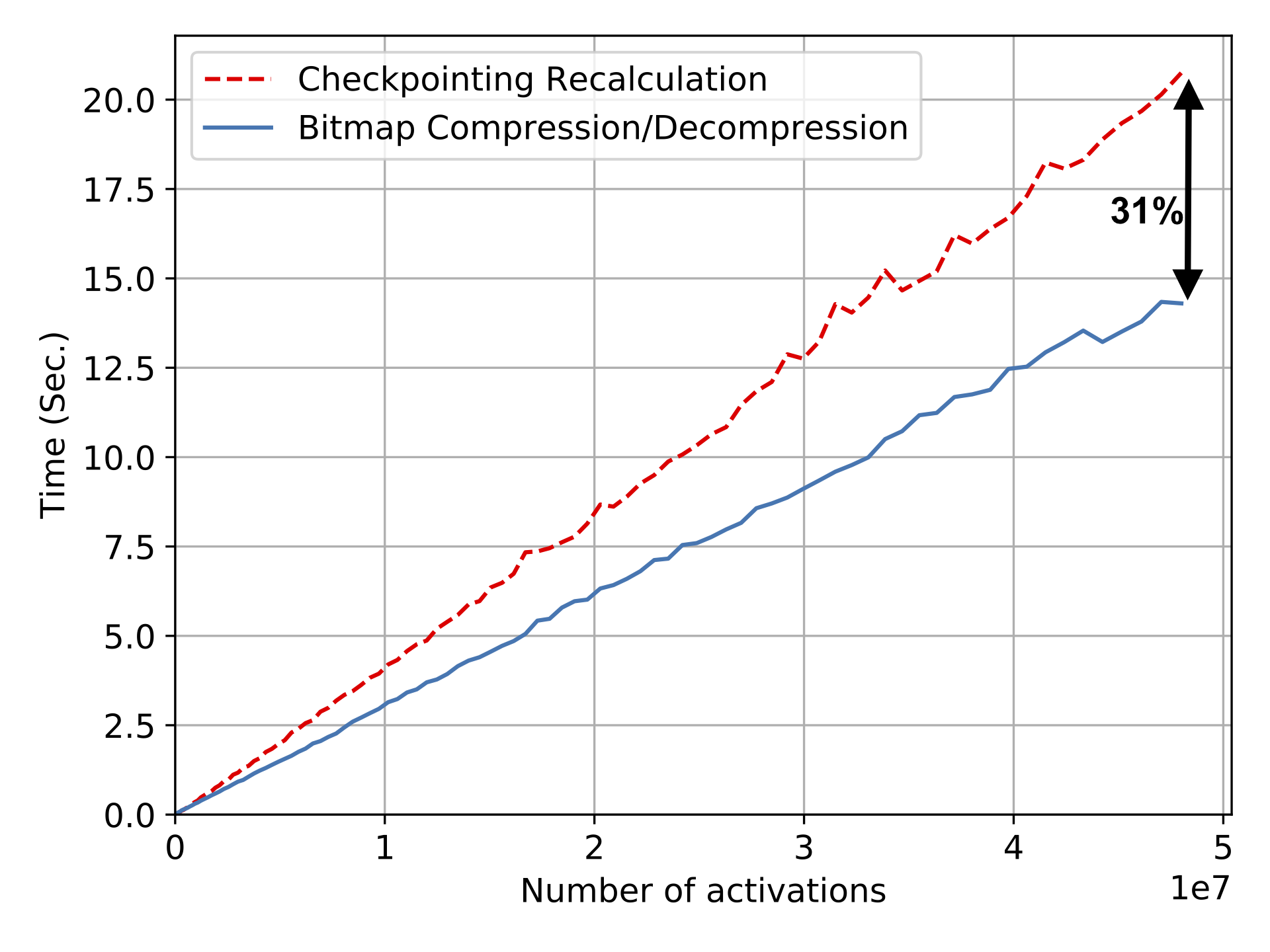}

    \caption{On-board (Jetson Nano) runtime reduction of bitmap compression/decompression vs. recalculating activations (checkpointing) at 50\% sparsity.}

    \label{fig:runtime}
\end{figure}

\textbf{Runtime.}
We analyzed the runtime savings that the memory operations (from bitmap compression and decompression) offer when compared to the expensive matrix multiplication operations for one layer convolution.
Figure \ref{fig:runtime} shows that the bitmap compression saves up to 31\% of runtime at the activation size of over 48m elements.
While memory operations are slower than floating-point operations (in terms of clock cycles), in the edge training use case, we find that it is indeed faster to perform pure memory operations than floating-point calculations due to the fact that memory is the constraint.
This is also desirable as it would reduce the total power consumed by the training.
Therefore, our bitmap compression/decompression method is even more compute-efficient.

\textbf{Summary.}
Using sparse bitmap compression is an efficient way to reduce the memory footprint for training deep learning models on the edge, with 18-53\% overall training memory reduction of well-established image classification models.
We have shown that memory reduction can indeed be measured empirically, achieving up to 34\% memory reduction in storing convolution activations at a sparsity level of 50\% (resulting from ReLU activations).
Our method is orthogonal to existing methods in the literature, and further pushes down the memory footprint.
For example, pruning increases sparsity to more than 75\%, which can save up to 56\% of the activations memory footprint, with negligible effect on the accuracy.
Furthermore, low-precision can also double down memory consumption with up to 55-75\% reduction.
In addition, using bitmap compression for saving the activation outperforms classic checkpointing by eliminating the need for reproducing expensive matrix operations.

%\section{Future Work}
%\label{sec_future_work}
%\input{7future_work}

\section{Conclusion and Future Work}
\label{sec_concolusion}

We propose BitTrain \--- a \textit{Sparse Bitmap Compression} technique for memory-efficient training on the edge.
Unlike previous methods that focus on saving memory to train deeper models on server-grade infrastructure, BitTrain directly optimizes the training memory footprint by addressing the most critical component of it \--- activations memory.
We exploit activations sparsity and save them in a compressed format that scales sub-linearly with the total number of activations.
BitTrain reduces the training memory footprint with no effect on the accuracy.
Extensive experiments on benchmark datasets show that our method is orthogonal to existing work, and can be efficiently combined with them.
BitTrain is a step further for efficient learning on the edge.

BitTrain is a first step towards enabling a new frontier in edge intelligence capabilities.
In the future, BitTrain can be extended to further enable on-device learning.
First, the compression and decompression processed can be integrated into the autograd libraries of the modern deep learning frameworks.
The idea is to provide a seamless implementation similar to the checkpointing API.
Second, defining native matrix operations on the bitmap format would make it more convenient for transfer learning on the edge.
This will push the frontier of special hardware support for model ``adapting'' pre-trained models to new data locally on constrained devices.
Third, combining all discussed methods in an integrated and efficient implementation would improve the empirical results, and push them closer to the theoretical calculations.
This work democratize AI for low-resource settings and can also advance the state-of-the-art of privacy-sensitive AI applications.

\noindent \textbf{Acknowledgements:} This work is partially supported by NSF grant 1814920 and DoD ARO grant W911NF-19-1-0484.

\bibliographystyle{ACM-Reference-Format}
\bibliography{refbib}

\end{document}